\pdfoutput=1

\documentclass[11pt]{article}

\usepackage[final]{acl}
\usepackage[utf8]{inputenc}

\usepackage{times}
\usepackage{latexsym}

\usepackage[T1]{fontenc}

\usepackage{fontawesome}
\DeclareUnicodeCharacter{1F99C}{\parrot}

\usepackage[utf8]{inputenc}   
\usepackage{textcomp}         
\usepackage{tipa}             

\DeclareUnicodeCharacter{26B1}{\texturn}
\usepackage[utf8]{inputenc}

\usepackage{microtype}

\usepackage{inconsolata}

\usepackage{float}

\usepackage{hyperref}

\usepackage{graphicx}

\usepackage{multirow}
\usepackage{amsmath}
\usepackage{booktabs}

\usepackage{amsfonts}
\usepackage{booktabs}
\usepackage{siunitx}

\usepackage[mathletters]{ucs}

\usepackage{xcolor, colortbl, pgf}
\usepackage{enumitem}

\usepackage{pgf}
\usepackage{colortbl}

\definecolor{color1}{HTML}{93003a}
\definecolor{color2}{HTML}{cf3759}
\definecolor{color3}{HTML}{f4777f}
\definecolor{color4}{HTML}{ffbcaf}
\definecolor{color5}{HTML}{ffffe0}
\definecolor{color6}{HTML}{a5d5d8}
\definecolor{color7}{HTML}{73a2c6}
\definecolor{color8}{HTML}{4771b2}
\definecolor{color9}{HTML}{00429d}

\usepackage{array}

\newcommand*{\opacity}{30}
\newcommand*{\minval}{0.00}
\newcommand*{\maxval}{1.0}

\newcommand{\gradient}[1]{
    \ifdimcomp{#1pt}{>}{\maxval pt}{#1}{
        \ifdimcomp{#1pt}{<}{\minval pt}{#1}{
            \pgfmathparse{int(round(8*(#1/(\maxval-\minval))-(\minval*(8/(\maxval-\minval)))))}
            \xdef\tempa{\pgfmathresult}
            \ifcase\tempa
                \cellcolor{color1!\opacity} #1\or
                \cellcolor{color2!\opacity} #1\or
                \cellcolor{color3!\opacity} #1\or
                \cellcolor{color4!\opacity} #1\or
                \cellcolor{color5!\opacity} #1\or
                \cellcolor{color6!\opacity} #1\or
                \cellcolor{color7!\opacity} #1\or
                \cellcolor{color8!\opacity} #1\or
                \cellcolor{color9!\opacity} #1
            \fi
    }}
}

\title{Sonnet or Not, Bot?\\Poetry Evaluation for Large Models and Datasets}

\newcommand{\uw}{$^\diamondsuit$}
\newcommand{\aitwo}{$^\spadesuit$}
\newcommand{\aspace}{\hspace{\fontdimen2\font}}

\author{Melanie Walsh\uw \aspace\aspace\aspace Anna Preus\uw \aspace\aspace\aspace Maria Antoniak\aitwo 
\vspace{.3em}\\
\small{\uw University of Washington \aspace \aspace \aitwo University of Copenhagen}}

\definecolor{maria-color}{HTML}{7881F2}

\definecolor{todo-color}{HTML}{ff3232}

\definecolor{light-color}{HTML}{A0A0A0}
\definecolor{lighter-color}{HTML}{727272}  

\begin{document}
\maketitle

\begin{abstract}
Large language models (LLMs) can now generate and recognize poetry. 
But what do LLMs really know about poetry? We develop a task to evaluate how well LLMs recognize one aspect of English-language poetry---poetic form---which captures many different poetic features, including rhyme scheme, meter, and word or line repetition. 
By using a benchmark dataset of over 4.1k human expert-annotated poems, we show that state-of-the-art LLMs can successfully identify both common and uncommon \textit{fixed} poetic forms---such as \textit{sonnets}, \textit{sestinas}, and \textit{pantoums}---with surprisingly high accuracy. 
However, performance varies significantly by poetic form; the models struggle to identify \textit{unfixed} poetic forms, especially those based on topic or visual features. 
We additionally measure how many poems from our benchmark dataset are present in popular pretraining datasets or memorized by GPT-4, finding that pretraining presence and memorization may improve performance on this task, but results are inconclusive.
We release a benchmark evaluation dataset with 1.4k public domain poems and form annotations, results of memorization experiments and data audits, and code.
\end{abstract}

\section{Introduction}

\begin{quote}
    \textit{Writing free verse is like playing tennis with the net down.}\\
- Robert Frost
\end{quote} \label{frost-quote}

\begin{figure}[t]
    \centering
    \includegraphics[width=\linewidth]{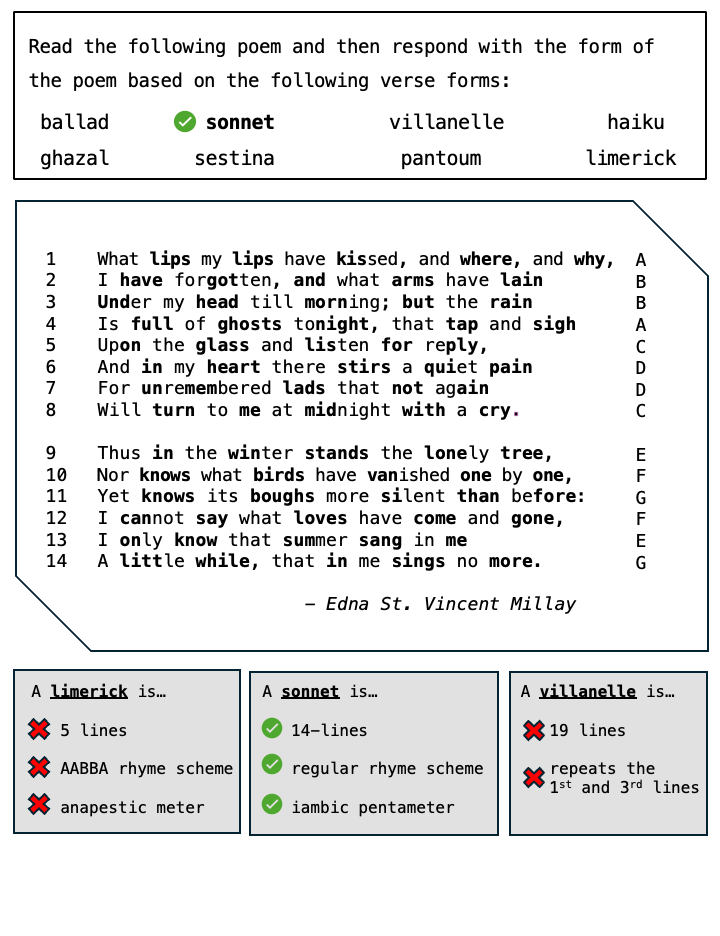}
    \vspace{-1.5cm}
    \caption{We develop a task to evaluate how well LLMs can identify \textit{poetic form}  for more than 20 poetic forms and formal elements in the English language. 
    This is a challenging task because poetic form is determined by a combination of factors: rhyme scheme, meter, repetition, number of lines, and/or subject matter. 
    }
    \label{figure:overview}
\end{figure}

The poetic capabilities of large language models (LLMs) have been cited prominently by journalists, social media users, and even LLM developers and marketers \cite{pogue_ai_2023v2, zahn_what_2022v2, roose_google_2024}. 
Google named its first chatbot ``Bard,'' a traditional term for a poet and the nickname of William Shakespeare, and Anthropic named two of its 2024 Claude models after popular poetic forms, ``Sonnet'' and ``Haiku.'' 
Microsoft released an ad that featured its Bing chatbot writing poetry \cite{bing_what_2023},
as well as an instruction guide for how to write poems with Copilot,
including a list of suggested forms to try \cite{microsoft_how_2023}.
Generated poetry was also one of the first LLM outputs to go viral on social media and remains popular there \cite{thomas_h_ptacek_tqbf_im_2022}.
Poetry is a lightning rod for the marketing and popular imagination of LLM capabilities as a signifier of human creativity and complexity, as well as a popular and culturally significant art form with a long history.

But what do LLMs really know about poetry? 
Prior natural language processing (NLP) research has focused on poetry generation \citep{zhang-lapata-2014-chinese, wockener_end--end_2021, chen_evaluating_2024}, poetry summarization \citep{mahbub-etal-2023-unveiling}, and detection of individual poetic forms \cite{abdibayev_automating_2021}. But we need broader evaluation of a wider range of poetic forms and features, and updated audits of LLMs. Poetry is valuable for auditing LLM capacities (even beyond the domain of literature) because it encompasses many challenging aspects of human language and expression. 
Poetry combines verbal, aural, and visual elements in unique configurations---that is, the substance, sound, and (in written poetry) appearance of words on the page (e.g. white space) all matter. 
Poetry also communicates deep emotion and meaning in non-literal, ambiguous ways, employing rhetorical devices that are difficult for LLMs such as figurative language, irony, and allusion \cite{jhamtani_investigating_2021, jang_figurative_2023}.

To measure LLMs' poetic capabilities, we develop a task to evaluate how well LLMs recognize more than 20 poetic forms and formal elements in the English language. 
Poetic form captures many different poetic features, including rhyme scheme, meter (the underlying beat), and word or line repetition (see \ref{poetic-form-definitions}).
\textit{Sonnets}, \textit{limericks}, and \textit{haiku} are well-known forms. But there are also less-known, more complicated forms like \textit{sestinas} (which repeat the same six endwords in an intricate pattern) or \textit{pantoums} (which repeat the second and fourth lines of stanzas in an alternating pattern).
To make matters more complicated, authors often play with or defy poetic form on purpose.
Thus, identifying poetic form is a difficult and divisive task even for human experts, as we show in a formative study.

We use this task to reflect on LLMs' current poetic capabilities, as well as the challenges and pitfalls of creating NLP benchmarks for poetry and other creative tasks. 
In particular, we use this task to audit and reflect on the poems included in popular pretraining datasets.
A complication is that the circulation of poetry is different from other literary texts, like fiction books and long-form prose, which have received the most attention in prior work.
Poems are often short and ``portable.'' On the web and within the publishing industry, individual poems can ``travel'' across multiple websites and anthologies in ways that previously studied books data \citep{chang-etal-2023-speak} do not, resulting in increased memorization issues that will affect any poetry evaluation benchmark.

We find that LLMs---particularly GPT-4 and GPT-4o---can successfully identify both common and uncommon \textit{fixed} poetic forms, such as \textit{sonnets}, \textit{sestinas}, and \textit{pantoums}, at surprisingly high accuracy levels when compared to human experts. 
But performance varies widely by poetic form and feature; the models struggle to identify \textit{unfixed} poetic forms, especially ones based on topic or visual features. 
When we compare model performance on poems from major online poetry institutions, popular pretraining datasets, and print books with little to no digital presence, we do not see major differences in classification performance. 

Our findings have implications for NLP studies of poetry/creative text generation and analysis, digital humanities and cultural analytics research, as well as cultural heritage collections, libraries, and archives that include poetry.

Our contributions include:
\begin{itemize}
\itemsep-.1em
    \item the introduction of the poetic form detection task, with a comparison to formative human study of poetry experts,
    \item a set of benchmark evaluation experiments using 4.1k poems,
    \item an analysis of poems found in popular pretraining data and memorized by models,
    \item code, data (1.4k public domain poems and form annotations), and metadata (pretraining inclusion and model memorization) that we release to the public.\footnote{\url{https://github.com/maria-antoniak/poetry-eval}} 
\end{itemize}

\section{Poetic Form} 
\label{poetic-form}

\paragraph{Subjective, Fluid, Context-Dependent.} Traditionally, ``form'' refers to ``the manner in which a poem is composed as distinct from what the poem is about,'' and it can also refer more broadly to ``genre or kind of composition''  \cite{preminger_princeton_2015}. Poetic form can be defined by particular patterns of sound, referred to as prosody, and/or by visual patterns. In scholarship on poetics, forms are fluid and sometimes overlapping. They exist within specific cultural and linguistic contexts, but also travel across them \cite{ramazani_transnational_2009}. They are socially and historically constructed and have been the subject of heated debates \cite{martin_rise_2012}, while also demonstrating remarkable durability across time (a number of the forms we test originated over 1,000 years ago). 
Since we focus on a corpus of mostly English-language poetry, the forms we focus on are all common in English, although most of them originated in other languages. For ``fixed'' forms, there are often specific rules and complex patterns of versification, but these rules are also likely to be stretched or broken by poets \cite{leighton_form_2008}. Like other literary genres, forms serve as ``frameworks of expectation'' \cite{seitel_theorizing_2003} that are called up and manipulated in meaningful ways by writers. This makes it inherently difficult and subjective to evaluate poetic form.

\paragraph{Fixed and Unfixed Forms.} We divide the poetic forms we consider into three categories: fixed forms, formal elements, and unfixed forms. See \ref{poetic-form-definitions}
for definitions and examples of all poetic forms.
\textbf{Fixed forms} follow particular patterns in terms of number of lines, meter, rhyme, and/or repetition. \textit{Sonnets} (14 lines, usually with a specific rhyme scheme and meter) and \textit{villanelles} (19 lines with specific repeating lines) are both fixed forms. \textbf{Formal elements} may be component parts of other forms or may define a poem as a whole. Common stanza types like \textit{quatrains} (a group of four lines) or meters like \textit{blank verse} (unrhymed lines with 10 alternating stressed and unstressed syllables) are formal elements. 
\textbf{Unfixed forms} are defined by particular subject matter or kinds of content, rather than by patterns of repetition or sound. These are forms like \textit{elegy} (writing about loss), 
which come in a variety of shapes, sizes, and patterns.

These categorizations are recognized as imperfect, and they are neither stable nor discrete. A type of poetry like \textit{haiku} has a common fixed form in English---three lines consisting of 5, 7, and 5 syllables---but \textit{haiku }can also refer to concise, non-narrative poems with any number of lines that tend to focus on natural imagery \cite{sato_haiku_2018}. 
We chart the specific poetic features---rhyme, repetition, meter, fixed topic, etc.---that are common for each form in Appendix \ref{appendix-poetic-features}, and we analyze classification performance by poetic feature in Table \ref{table:poetic_features}. But it is important to note that the same poetic features can apply to different forms in different ways. For example, \textit{blank verse} and \textit{free verse} both hinge on meter (underlying beat), but blank verse is typically written in a specific meter called iambic pentameter (10 alternating stressed and unstressed syllables with a musical quality) while free verse does not have any consistent meter at all. 
Lastly, a single poem can also belong to more than one category. For example, John Keats's ``Ode on a Grecian Urn'' is an \textit{ode}, but it is also an example of \textit{ekphrasis} (writing about art), since it describes a decorated vase. 
To manage this complexity, we exclude poems with multiple relevant tags in the same category, such as \textit{pastoral} and \textit{elegy} (both unfixed forms). 
We believe that multi-label classification is an important avenue for future work.

\paragraph{Meta-Discussion of Poetic Form.} Like Keats in ``Ode on a Grecian Urn,'' many authors include the name of the form they are engaged with in the title or text of a poem. 
While in the context of NLP evaluation these explicit mentions of a poem's form may seem to ``give away'' the correct answer, they are a fundamental aspect of poetry and are integral to a human reading experience. 
To address this issue, we experiment with prompts where the poem title both is and is not included, and we incorporate basic statistics about how many poems explicitly name the form in our results (see Figures \ref{fig:fixed-form-performance-prompt}, \ref{fig:unfixed-form-performance-prompt}).

\section{Data}

We curate more than 4.1k poems, mostly in the English language, which were categorized with their poetic forms by human annotators, and either published online or collected in books.

\subsection{Poetry Sources}

\paragraph{Poetry Foundation.} Poetry Foundation is a non-profit that works ``to amplify poetry and celebrate poets'' \cite{foundation_about_2024}. 
The organization runs \textit{Poetry} magazine, and it also hosts an online database\footnote{\url{https://www.poetryfoundation.org/}} of English-language poetry with more than 47k poems.

\paragraph{Academy of American Poets.} The Academy of American Poets is also a non-profit whose mission is ``to support American poets at all stages of their careers and to foster the appreciation of contemporary poetry'' \cite{poets_about_2024}. The organization hosts a website\footnote{\url{https://poets.org/}} that includes more than 10k poems. 

\paragraph{Manually Digitized Poetry Books.} We also manually digitize a range of poetry collections and anthologies organized by form that, when searched in the international library database WorldCat, did not have obvious e-books or presences in major databases (e.g. HathiTrust Digital Library). See \ref{manually-digitized-bib} for full list of books.

\vspace{0.2cm}
\noindent
To our knowledge, the collections from the Poetry Foundation and Academy of American Poets represent the largest collections of poetry with human-labeled forms that extend into the present day.
They are both well-respected poetry institutions with significant engagement from poets and poetry scholars. 
Both institutions have taken great care in formatting their poems with correct white space and line breaks in the HTML of their websites---an aspect of the poems that is essential to understanding both their form and meaning.

\subsection{Poetry Curation and Processing}

\begin{table}[t]
    \centering
    \scriptsize
    \begin{tabular}{@{}m{1.5cm}|m{0.7cm}m{.7cm}m{0.3cm}m{0.9cm}m{1.3cm}@{}}
    \toprule
    \textbf{Poetic Form x Source} & \tiny{Poetry\newline Foundation} & \tiny{Academy of\newline American\newline Poets} & \tiny{Both} & \tiny{Manually\newline Digitized} & \tiny{\textbf{Total}} \\
    \midrule
    \multicolumn{6}{@{}l}{\textit{Fixed Forms}} \\
    Ballad & 96 & 12 & 2 & 25 & 135 \\
    Ghazal & 21 & 19 & 0 & 40 & 80 \\
    Haiku & 25 & 24 & 1 & 42 & 92 \\
    Limerick & 6 & 1 & 0 & 42 & 49 \\
    Pantoum & 11 & 14 & 0 & 42 & 67 \\
    Sestina & 16 & 23 & 2 & 40 & 81 \\
    Sonnet & 376 & 467 & 13 & 40 & 896 \\
    Villanelle & 43 & 17 & 3 & 40 & 103 \\
    \midrule
    \multicolumn{6}{@{}l}{\textit{Formal Elements}} \\
    Blank Verse & 209 & 0 & 0 & 0 & 209 \\
    Free Verse & 387 & 0 & 0 & 0 & 387 \\
    Common\newline Measure & 112 & 0 & 0 & 0 & 112 \\
    Couplet & 398 & 0 & 0 & 0 & 398 \\
    Quatrain & 89 & 0 & 0 & 0 & 89 \\
    Tercet & 94 & 0 & 0 & 0 & 94 \\
    \midrule
    \multicolumn{6}{@{}l}{\textit{Unfixed Forms}} \\
    Ars Poetica & 23 & 68 & 3 & 0 & 94 \\
    Aubade & 11 & 5 & 0 & 0 & 16 \\
    Concrete\newline Poetry & 24 & 0 & 0 & 0 & 24 \\
    Dramatic\newline Monologue & 158 & 32 & 1 & 0 & 191 \\
    Ekphrasis & 81 & 63 & 1 & 0 & 145 \\
    Elegy & 193 & 59 & 2 & 10 & 264 \\
    Ode & 73 & 43 & 3 & 2 & 121 \\
    Pastoral & 75 & 0 & 0 & 0 & 75 \\
    Prose Poem & 334 & 141 & 0 & 0 & 475 \\
    \midrule
    \textbf{Total} & \textbf{2,855} & \textbf{988} & \textbf{31} & \textbf{323} & \textbf{4,197 poem/form pairs} \\
    \bottomrule
    \end{tabular}
    \caption{The distribution of poems in our collected dataset by form and source.}
    \label{table:poem-distribution}
\end{table}

We select  poems in the following categories delineated by the Poetry Foundation on their website: verse forms, stanza forms, meters, and types/modes. 
Conceptually, as discussed in \S\ref{poetic-form}, we frame these tag categories as \textbf{fixed forms}, \textbf{formal elements}, and \textbf{unfixed forms} (see Table \ref{table:poem-distribution}).
The Academy of American Poets does not tag poems by meter or stanza form, so for these forms, we only use the Poetry Foundation as our source.

We scrape up to 400 poems per available form on each of the two websites. We exclude poems that have multiple relevant tags in the same form category, but we allow poems that may have multiple relevant tags in different form categories, such as \textit{blank verse} (formal element) and \textit{elegy} (unfixed form).
We preserve white space and line breaks in our dataset and see this as a central contribution. 

Additionally, we digitize 15 print poetry anthologies and collections tagged with each of the fixed forms that we consider, according to Library of Congress subject headings via WorldCat.

We release 1.4k public domain poems from this dataset with form annotations, as well as other accompanying metadata, such as subject tags and author birth and death years when available. 
We do not make in-copyright poems available.

\section{Auditing Pretraining Data for Poems}
\label{section:pretraining-datasets}

Online resources like Poetry Foundation are valuable in large part because they make thousands of poems available on the internet for free. 
However, this also means that these specific poems are more likely to be present in the training data of LLMs, leading to memorization issues that could affect performance on our form classification task. 
Prior work has found significant amounts of poetry memorization in large models like GPT-3.5 \citep{dsouza_chatbot_2023}.
We perform experiments to probe pretraining datasets for our poems, relying on both prompt-based model probes and direct searches of released pretraining data.

To directly search for poem texts, we rely on the \textbf{Dolma} open pretraining dataset \citep{soldaini2024dolma}.
Dolma is a ``three-trillion-token English corpus, built from
a diverse mixture of web content, scientific papers, code, public-domain books, social media, and encyclopedic materials.''
We query the Dolma dataset using the \textsc{What's In My Big Data} (\textsc{WIMBD}) platform \citep{elazar2023whats},\footnote{\url{https://github.com/allenai/wimbd}} which allows us to search for exact strings and returns all matches along with their associated metadata, including the data source, the original web domain, the surrounding text, and other information.
We split each poem into lines, and we remove lines with fewer than four whitespace-delimited tokens to avoid unspecific matches. 
We truncate lines at 20 tokens for query efficiency.
We release our search results publicly.

\begin{table}[t]
    \centering
    \scriptsize
    \begin{tabular}{@{}p{2.6cm}|p{1cm}p{0.9cm}p{1.8cm}@{}}
    \toprule
    \textbf{Domain} & \textbf{$N$ Poems} & \textbf{$N$ Lines} & \textbf{Domain Type} \\
    \midrule
    \texttt{github.com}
    & 740
    & 40,724
    & \textit{content hosting}
    \\[0.5ex]
    \texttt{reddit.com}
    & 733
    & 9,773
    & \textit{social media}
    \\[0.5ex]
    \texttt{books.google}
    & 545 	  	 
    & 113,373
    & \textit{books}
    \\[0.5ex]
    \texttt{engpoetry.com}
    & 477 	  	 
    & 4,923
    & \textit{\textbf{poetry}}
    \\[0.5ex]
    \texttt{gutenberg.org}
    & 431 	  	 
    & 15,363
    & \textit{books}
    \\[0.5ex]
    \textbf{\texttt{poets.org}}
    & 256 	  	 
    & 2,290
    & \textit{\textbf{poetry}}
    \\[0.5ex]
    \texttt{poemhunter.com}
    & 243 	  
    & 1,589
    & \textit{\textbf{poetry}}
    \\[0.5ex]
    \texttt{quotes.yourdiction...}
    & 217 	  	 
    & 2,611
    & \textit{quotes}
    \\[0.5ex]
    \texttt{enotes.com}
    & 200 	  	
    & 872
    & \textit{study guides}
    \\[0.5ex]
    \texttt{poetryexplorer.net}
    & 181 	  
    & 649
    & \textit{\textbf{poetry}}
    \\[0.5ex]
    \texttt{poetrysoup.com}
    & 179 	 
    & 3,126 
    & \textit{\textbf{poetry}}
    \\[0.5ex]
    \texttt{inspirationalstori...}
    & 171
    & 866
    & \textit{stories}
    \\[0.5ex]
    \texttt{free-translator.com}
    & 147 	 
    & 2,555
    & \textit{translation}
    \\[0.5ex]
    \texttt{hotfreebooks.com}
    & 145 	  	
    & 2,110
    & \textit{books}
    \\[0.5ex]
    \texttt{m.poemhunter.com}
    & 142 	  	 
    & 1,218
    & \textit{\textbf{poetry}}
    \\[0.5ex]
    \texttt{rpo.library.utoron...}
    & 132 	  	 
    & 1,026
    & \textit{books}
    \\[0.5ex]
    \texttt{poemine.com}
    & 129 	  	
    & 835
    & \textit{\textbf{poetry}}
    \\[0.5ex]
    \texttt{semanticscholar.org}
    & 127 	  	 
    & 442
    & \textit{academic papers}
    \\[0.5ex]
    \texttt{internetpoem.com}
    & 125 	  	 
    & 798
    & \textit{\textbf{poetry}}
    \\[0.5ex]
    \texttt{azquotes.com}
    & 121 	  	
    & 460
    & \textit{quotes}
    \\[0.5ex]
    \bottomrule
    \end{tabular}
    \caption{The source domains with the highest number of detected poems. } 
    \label{table:domains-from-dolma}
\end{table}

\subsection{How many poems are in pretraining data?}

\begin{figure}[t]
    \centering
        \centering
        \includegraphics[width=\linewidth]{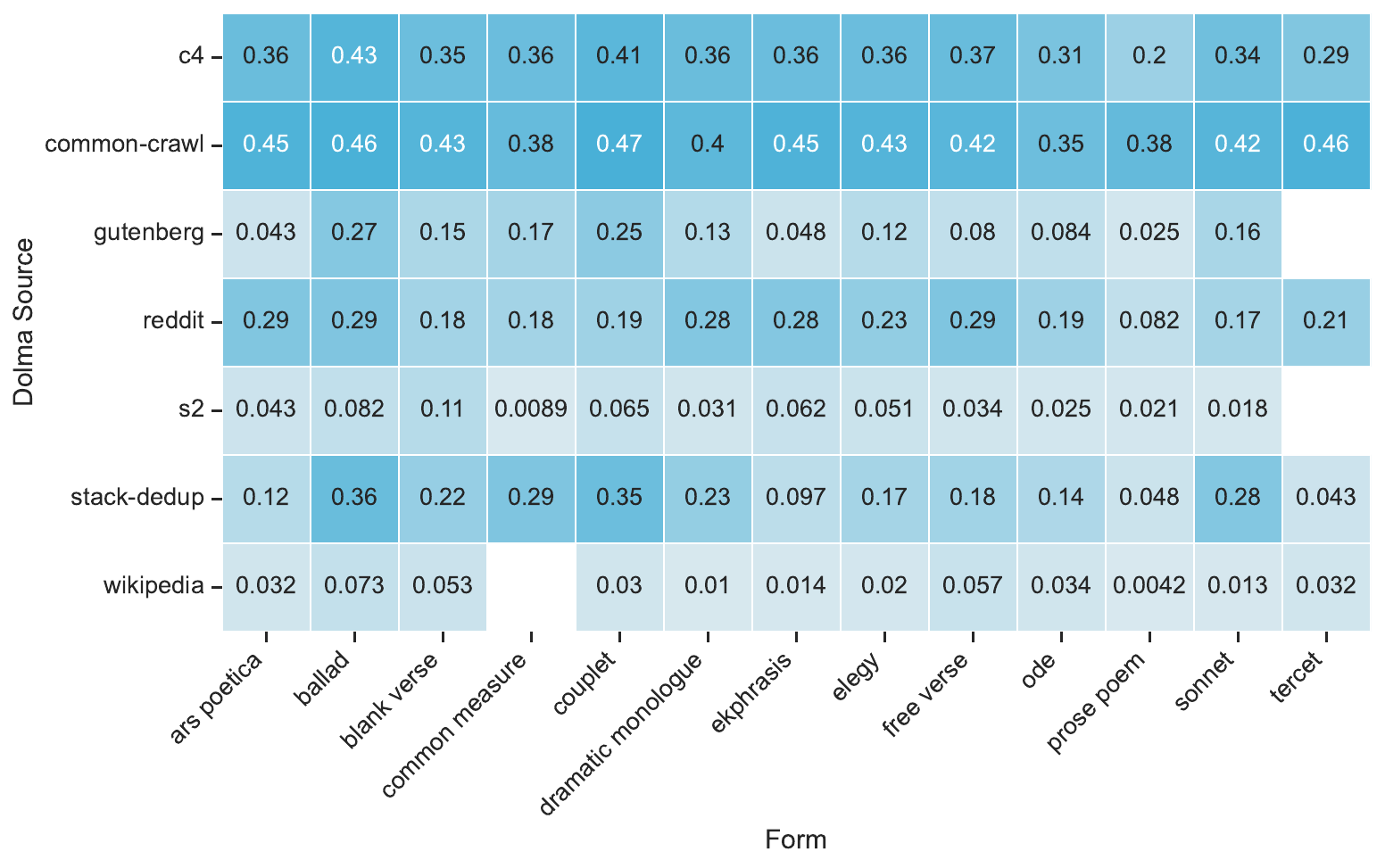}
    \caption{The proportion of all poems for a given form that were detected (at least one line) in the source data for Dolma. We include only the most frequent forms. Poems can appear in multiple sources and belong to muultiple forms. The Common Crawl dataset dominates, and some sources like Project Gutenberg contain significant percentages of only certain forms like \textit{ballads} and \textit{ couplets}.}
    \label{figure:heatmap-forms-sources}
\end{figure}

\begin{figure}[t]
    \centering
        \centering
        \includegraphics[width=0.7\linewidth]{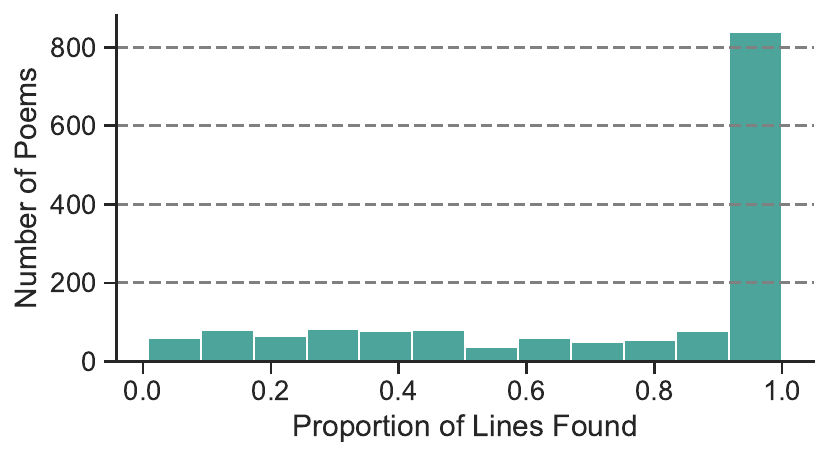}
    \caption{The proportions of lines detected in Dolma per poem (only those with at least one line detected). If at least one line from a poem is detected, it is likely that all the lines will be detected somewhere in Dolma.}
    \label{figure:histogram-lines-found}
\end{figure}

We find that about half of the poems (57\%) are \textit{not} present in Dolma (not even one line is detected).
This does not guarantee that these poems are not present in the pretraining data for industry models whose pretraining data is not disclosed and likely includes many in-copyright texts---but this provides us with one publicly available clue.
Fig. \ref{figure:heatmap-forms-sources} shows the forms and the proportions of their associated poems that were detected in Dolma, categorized by the Dolma source.
About 30\% of our poems are found in the Common Crawl data included in Dolma, with the C4 dataset close behind.
Wikipedia and Semantic Scholar contain the fewest detected poems.
Overall, if at least one line from a poem is detected, it is likely that all the lines will be detected somewhere in Dolma (see Figure \ref{figure:histogram-lines-found}).

\subsection{Where does this poetry data come from?}
\label{memorization}

Examining the web domains from which the Dolma data was sourced, we find that large websites like Github, Reddit, and Google Books dominate the rankings (Table \ref{table:domains-from-dolma}).
Many poetry-specific websites like \url{engpoetry.com} and \url{poets.org} (the website of the Academy of American Poets, one of our data sources) are also present in the top ranked domains, as are domains related to books.
Figure \ref{figure:heatmap-forms-sources} shows the distribution across data sources, with the Common Crawl dataset dominating, and some sources, such as Project Gutenberg, containing significant percentages only for certain forms like \textit{ballads} and \textit{ couplets}.
Models trained on different mixes of these sources could be more or less capable of recognizing certain forms.

\subsection{Are these poems memorized?}
We additionally replicate the tests from \citet{dsouza_chatbot_2023} by prompting GPT-4 to produce the next five lines of a poem, given its title, author, and first line (see \ref{appendix-memorization-prompt} for our prompt).
We then check for any overlapping five-gram span between the model's output and the original poem text; hand-annotations for 300 random poems indicate that this is a viable method to check for memorization (97\% accuracy).
We find that 41\% of poems are memorized by GPT-4, and 46\% of these memorized poems are also found in Dolma.
This indicates that more poetry data is available in the training of closed models like GPT-4 than is available in Dolma---perhaps within datasets of published books that are not included in Dolma---and memorization is an issue that can be partly but not fully addressed by current open resources.

\section{Establishing Difficulty of Task}

To establish the difficulty and significance of the poetic form classification task, we conduct a small formative study\footnote{
``Formative'' studies, typically small in scope, help inform a project or its development but do not ``empirically test hypotheses'' \cite{buffardi_formative_2014, qian_formative_2020}.} with human experts, and we also test a fine-tuned LLM and a supervised machine learning classifier.
Through these initial experiments, we confirm that this task cannot be accomplished easily by human experts, by basic fine-tuned models, or by traditional lexical approaches.

\subsection{Human Expert Study Design}

We conduct a small formative survey with 15 literature and poetry scholars, asking them to categorize four example poems from our dataset based on the text alone.\footnote{The survey asked respondents to answer whether they considered themselves ``a literary scholar`` (15 respondents) and ``a poetry scholar, specifically'' (8 respondents). We only consider results from literary/poetry scholars.} We select four poems that range in difficulty, conventionality, and ambiguity. We select three poems in common English-language poetic forms (\textit{sonnet}, \textit{ballad}, \textit{haiku}) and one poem in an uncommon form (\textit{pantoum}) (see Appendix \ref{formative-study} for poems).
We shared the survey in early 2024 on social media, with
colleagues, and to scholars associated with the literary studies conference Modern Language Association (MLA). 

\subsection{Human Expert Study Results}

Though most literary/poetry scholars correctly answered \textit{sonnet} and \textit{ballad} for Poems \href{https://www.poetryfoundation.org/poems/49146/piazza-piece}{\textbf{1}} and \href{https://www.poetryfoundation.org/poems/43773/prospice}{\textbf{2}} (67\% and 53\%), it was not an overwhelming majority. Answers were split between a wide variety of poetic forms (see Figure \ref{fig:scholar-study}). This split showcases the difficulty of the task even with relatively straightforward poems. 
Results for the two less common and conventional poems are even more interesting. While the majority of experts did \textit{not} identify the tagged form of Poems \href{https://www.poetryfoundation.org/poems/56355/my-brother-at-3-am}{\textbf{3}} and \href{https://www.poetryfoundation.org/poetrymagazine/poems/57528/poem-written-with-buson-in-a-minute}{\textbf{4}}, most LLMs did. All models except GPT-4o correctly identified Matthew Rohrer's \textit{haiku} based on the text alone, even though it is atypically long and does not follow the common 5-7-5 syllabic meter. GPT-4, GPT-4o, and Llama3 also correctly identified Natalie Diaz's \textit{pantoum} even though Diaz uses an approximate, rather than exact, repetition of lines. 

\subsection{Baseline Systems}

To further establish the difficulty of this task, we test both a fine-tuned RoBERTa classifier \citep{liu_roberta_2019}  and a traditional SVMs classifier with TF-IDF-weighted unigram features.
We randomly sample 100 poems from each of the most frequent forms, removing rare forms that do not have enough poems for reasonable fine-tuning and evaluation.
We then perform cross validation with five splits.
We find low performance across the forms, with a high of $0.81$ (F1) for \textit{prose poems} and $0.58$ (F1) for \textit{sonnets} and a low of $0.05$ (F1) for \textit{odes} and $0.06$ for \textit{elegies} (Table \ref{table:baseline-performance} in Appendix \ref{appendix-baselines}).

\section{Poetic Form Classification}

\subsection{Prompting Methods}

\begin{table*}[t]
    \scriptsize
    \centering
    \resizebox{1\linewidth}{!}{
    \begin{tabular}{lcccccccccccccccccccc}
    \toprule
    & \multicolumn{3}{c}{{\hyperref[sonnet]{Sonnet}}
} & \multicolumn{3}{c}{{\hyperref[limerick]{Limerick}}
} & \multicolumn{3}{c}{{\hyperref[haiku]{Haiku}}
} & \multicolumn{3}{c}{{\hyperref[ballad]{Ballad}}
} \\\cmidrule(lr){2-4}\cmidrule(lr){5-7}\cmidrule(lr){8-10}\cmidrule(lr){11-13}
               model & f1 & precision & recall & f1 & precision & recall & f1 & precision & recall & f1 & precision & recall \\\midrule
    GPT-3.5 & $\gradient{0.92}$\color{lighter-color}\tiny{$\pm .01$} & 0.94 \color{light-color}\tiny{$\pm .01$} & 0.91 \color{light-color}\tiny{$\pm .01$} & $\gradient{1.00}$\color{lighter-color}\tiny{$\pm .00$} & 1.00 \color{light-color}\tiny{$\pm .00$} & 1.00 \color{light-color}\tiny{$\pm .00$} & $\gradient{0.90}$ \color{lighter-color}\tiny{$\pm .03$} & 0.90 \color{light-color}\tiny{$\pm .04$} & 0.90 \color{light-color}\tiny{$\pm .05$} & $\gradient{0.78}$ \color{lighter-color}\tiny{$\pm .04$} & 0.82 \color{light-color}\tiny{$\pm .05$} & 0.75 \color{light-color}\tiny{$\pm .04$} \\
    GPT-4   & $\gradient{0.94}$\color{lighter-color}\tiny{$\pm .01$} & 0.98 \color{light-color}\tiny{$\pm .00$} & 0.90 \color{light-color}\tiny{$\pm .01$} & $\gradient{0.88}$\color{lighter-color}\tiny{$\pm .08$} & 0.78 \color{light-color}\tiny{$\pm .14$} & 1.00 \color{light-color}\tiny{$\pm .00$} & $\gradient{0.97}$ \color{lighter-color}\tiny{$\pm .02$} & 0.98 \color{light-color}\tiny{$\pm .02$} & 0.96 \color{light-color}\tiny{$\pm .03$} & $\gradient{0.83}$ \color{lighter-color}\tiny{$\pm .03$} & 0.78 \color{light-color}\tiny{$\pm .04$} & 0.88 \color{light-color}\tiny{$\pm .04$} \\
    GPT-4o  & $\gradient{0.94}$\color{lighter-color}\tiny{$\pm .01$} & 0.99 \color{light-color}\tiny{$\pm .00$} & 0.89 \color{light-color}\tiny{$\pm .01$} & $\gradient{0.93}$\color{lighter-color}\tiny{$\pm .05$} & 0.88 \color{light-color}\tiny{$\pm .08$} & 1.00 \color{light-color}\tiny{$\pm .00$ }& $\gradient{0.90}$ \color{lighter-color}\tiny{$\pm .03$} & 0.93 \color{light-color}\tiny{$\pm .03$} & 0.86 \color{light-color}\tiny{$\pm .05$} & $\gradient{0.86}$ \color{lighter-color}\tiny{$\pm .02$} & 0.88 \color{light-color}\tiny{$\pm .03$} & 0.84 \color{light-color}\tiny{$\pm .03$} \\
    Claude  & $\gradient{0.95}$\color{lighter-color}\tiny{$\pm .01$} & 0.95 \color{light-color}\tiny{$\pm .01$} & 0.95 \color{light-color}\tiny{$\pm .01$} & $\gradient{0.88}$\color{lighter-color}\tiny{$\pm .08$} & 0.78 \color{light-color}\tiny{$\pm .12$} & 1.00 \color{light-color}\tiny{$\pm .00$} & $\gradient{0.93}$ \color{lighter-color}\tiny{$\pm .04$} & 0.98 \color{light-color}\tiny{$\pm .02$} & 0.88 \color{light-color}\tiny{$\pm .05$} & $\gradient{0.78}$ \color{lighter-color}\tiny{$\pm .03$} & 0.94 \color{light-color}\tiny{$\pm .03$} & 0.66 \color{light-color}\tiny{$\pm .03$} \\
    Mixtral & $\gradient{0.92}$\color{lighter-color}\tiny{$\pm .01$} & 0.96 \color{light-color}\tiny{$\pm .00$} & 0.89 \color{light-color}\tiny{$\pm .01$} & $\gradient{0.88}$\color{lighter-color}\tiny{$\pm .08$} & 0.78 \color{light-color}\tiny{$\pm .11$} & 1.00 \color{light-color}\tiny{$\pm .00$} & $\gradient{0.79}$ \color{lighter-color}\tiny{$\pm .03$}  & 0.94 \color{light-color}\tiny{$\pm .03$} & 0.68 \color{light-color}\tiny{$\pm .05$} & $\gradient{0.74}$ \color{lighter-color}\tiny{$\pm .03$} & 0.72 \color{light-color}\tiny{$\pm .04$} & 0.75 \color{light-color}\tiny{$\pm .04$} \\
    Llama3  & $\gradient{0.73}$\color{lighter-color}\tiny{$\pm .01$} & 1.00 \color{light-color}\tiny{$\pm .00$} & 0.58 \color{light-color}\tiny{$\pm .01$} & $\gradient{0.70}$\color{lighter-color}\tiny{$\pm .08$} & 0.54 \color{light-color}\tiny{$\pm .09$} & 1.00 \color{light-color}\tiny{$\pm .00$} & $\gradient{0.79}$ \color{lighter-color}\tiny{$\pm .04$}  & 0.94 \color{light-color}\tiny{$\pm .04$} & 0.68 \color{light-color}\tiny{$\pm .06$} & $\gradient{0.45}$ \color{light-color}\tiny{$\pm .02$} & 0.31 \color{light-color}\tiny{$\pm .02$} & 0.80 \color{light-color}\tiny{$\pm .05$} \\\midrule
    & \multicolumn{3}{c}{{\hyperref[sestina]{Sestina}}
} & \multicolumn{3}{c}{{\hyperref[villanelle]{Villanelle}}
} & \multicolumn{3}{c}{{\hyperref[pantoum]{Pantoum}}
} & \multicolumn{3}{c}{{\hyperref[ghazal]{Ghazal}}
}\\\cmidrule(lr){2-4}\cmidrule(lr){5-7}\cmidrule(lr){8-10}\cmidrule(lr){11-13}
               model & f1 & precision & recall & f1 & precision & recall & f1 & precision & recall & f1 & precision & recall \\\midrule
    GPT-3.5 & $\gradient{0.17}$ \color{lighter-color}\tiny{$\pm .07$} & 0.80 \color{light-color}\tiny{$\pm .02$} & 0.10 \color{light-color}\tiny{$\pm .04$} & $\gradient{0.62}$ \color{lighter-color}\tiny{$\pm .03$} & 0.46 \color{light-color}\tiny{$\pm .03$} & 0.94 \color{light-color}\tiny{$\pm .03$} & $\gradient{0.20}$ \color{lighter-color}\tiny{$\pm .10$} & 0.60 \color{light-color}\tiny{$\pm .22$} & 0.12 \color{light-color}\tiny{$\pm .07$} & $\gradient{0.57}$ \color{lighter-color}\tiny{$\pm .07$}  & 0.51 \color{light-color}\tiny{$\pm .08$} & 0.65 \color{light-color}\tiny{$\pm .07$} \\
    GPT-4   & $\gradient{0.87}$ \color{lighter-color}\tiny{$\pm .04$} & 0.86 \color{light-color}\tiny{$\pm .05$} & 0.88 \color{light-color}\tiny{$\pm .06$} & $\gradient{0.93}$ \color{lighter-color}\tiny{$\pm .03$} & 0.98 \color{light-color}\tiny{$\pm .02$} & 0.89 \color{light-color}\tiny{$\pm .05$} & $\gradient{0.81}$ \color{lighter-color}\tiny{$\pm .05$} & 0.71 \color{light-color}\tiny{$\pm .07$} & 0.96 \color{light-color}\tiny{$\pm .04$} & $\gradient{0.65}$ \color{lighter-color}\tiny{$\pm .03$} & 0.51 \color{light-color}\tiny{$\pm .03$} & 0.92 \color{light-color}\tiny{$\pm .04$} \\
    GPT-4o  & $\gradient{0.73}$ \color{lighter-color}\tiny{$\pm .04$} & 0.63 \color{light-color}\tiny{$\pm .05$} & 0.88 \color{light-color}\tiny{$\pm .05$} & $\gradient{0.92}$ \color{lighter-color}\tiny{$\pm .03$} & 0.90 \color{light-color}\tiny{$\pm .04$} & 0.95 \color{light-color}\tiny{$\pm .03$} & $\gradient{0.82}$ \color{lighter-color}\tiny{$\pm .06$} & 0.74 \color{light-color}\tiny{$\pm .07$} & 0.92 \color{light-color}\tiny{$\pm .07$} & $\gradient{0.51}$ \color{lighter-color}\tiny{$\pm .04$} & 0.36 \color{light-color}\tiny{$\pm .03$} & 0.90 \color{light-color}\tiny{$\pm .05$} \\
    Claude  & $\gradient{0.41}$ \color{lighter-color}\tiny{$\pm .05$} & 0.31 \color{light-color}\tiny{$\pm .04$} & 0.61 \color{light-color}\tiny{$\pm .08$} & $\gradient{0.58}$ \color{lighter-color}\tiny{$\pm .03$} & 0.58 \color{light-color}\tiny{$\pm .04$} & 0.57 \color{light-color}\tiny{$\pm .06$} & $\gradient{0.53}$ \color{lighter-color}\tiny{$\pm .06$} & 0.54 \color{light-color}\tiny{$\pm .08$} & 0.52 \color{light-color}\tiny{$\pm .08$} & $\gradient{0.54}$ \color{lighter-color}\tiny{$\pm .07$} & 0.59 \color{light-color}\tiny{$\pm .06$} & 0.50 \color{light-color}\tiny{$\pm .08$} \\
    Mixtral & $\gradient{0.26}$ \color{lighter-color}\tiny{$\pm .05$} & 1.00 \color{light-color}\tiny{$\pm .00$} & 0.15 \color{light-color}\tiny{$\pm .06$} & $\gradient{0.69}$ \color{lighter-color}\tiny{$\pm .05$} & 0.78 \color{light-color}\tiny{$\pm .05$} & 0.62 \color{light-color}\tiny{$\pm .07$} & $\gradient{0.56}$ \color{lighter-color}\tiny{$\pm .07$} & 0.47 \color{light-color}\tiny{$\pm .08$} & 0.68 \color{light-color}\tiny{$\pm .08$} & $\gradient{0.65}$ \color{lighter-color}\tiny{$\pm .04$} & 0.58 \color{light-color}\tiny{$\pm .05$} & 0.75 \color{light-color}\tiny{$\pm .05$} \\
    Llama3  & $\gradient{0.17}$ \color{lighter-color}\tiny{$\pm .07$} & 0.67 \color{light-color}\tiny{$\pm .19$} & 0.10 \color{light-color}\tiny{$\pm .04$} & $\gradient{0.32}$ \color{lighter-color}\tiny{$\pm .01$} & 0.20 \color{light-color}\tiny{$\pm .01$} & 0.87 \color{light-color}\tiny{$\pm .04$} & $\gradient{0.46}$ \color{lighter-color}\tiny{$\pm .05$} & 0.33 \color{light-color}\tiny{$\pm .04$} & 0.76 \color{light-color}\tiny{$\pm .08$} & $\gradient{0.27}$ \color{lighter-color}\tiny{$\pm .09$} & 0.40 \color{light-color}\tiny{$\pm .08$} & 0.20 \color{light-color}\tiny{$\pm .08$} \\\bottomrule
    \end{tabular}
    }
    \caption{LLM performance by model for the \textbf{fixed forms}, where the prompt includes only the poem text. Standard deviations are shown for 20 bootstrapped samples of poems.}
    \label{table:evaluation-results}
\end{table*}

\begin{table*}[t]
    \scriptsize
    \centering
    \resizebox{1\linewidth}{!}{
    \begin{tabular}{l@{\hspace{1.5em}}cc@{\hspace{2.5em}}cc@{\hspace{2.5em}}cc@{\hspace{2.5em}}cc@{\hspace{2.5em}}cc@{\hspace{2.5em}}cc}
    \toprule
    & \multicolumn{2}{c}{{\hyperref[sonnet]{Sonnet}}
} & \multicolumn{2}{c}{{\hyperref[prose]{Prose Poem}}
} & \multicolumn{2}{c}{{\hyperref[couplet]{Couplet}}} & \multicolumn{2}{c}{{\hyperref[free-verse]{Free Verse}}
} \\\cmidrule(lr){2-3}\cmidrule(lr){4-5}\cmidrule(lr){6-7}\cmidrule(lr){8-9}
               model & Memorized & Not Memorized & Memorized & Not Memorized & Memorized & Not Memorized & Memorized & Not Memorized \\\midrule
    GPT-3.5 & $\gradient{0.97}$ \color{lighter-color}\tiny{$\pm .01$} & $\gradient{0.55}$ \color{lighter-color}\tiny{$\pm .02$} & $\gradient{0.31}$ \color{lighter-color}\tiny{$\pm .06$} & $\gradient{0.55}$ \color{lighter-color}\tiny{$\pm .03$}  & $\gradient{0.14}$ \color{lighter-color}\tiny{$\pm .03$} & $\gradient{0.19}$ \color{lighter-color}\tiny{$\pm .05$} & $\gradient{0.90}$ \color{lighter-color}\tiny{$\pm .02$} & $\gradient{0.80}$ \color{lighter-color}\tiny{$\pm .01$} \\ 
    GPT-4   & $\gradient{0.98}$ \color{lighter-color}\tiny{$\pm .01$} & $\gradient{0.88}$ \color{lighter-color}\tiny{$\pm .01$} & $\gradient{0.79}$ \color{lighter-color}\tiny{$\pm .05$} & $\gradient{0.87}$ \color{lighter-color}\tiny{$\pm .01$}  & $\gradient{0.52}$ \color{lighter-color}\tiny{$\pm .04$} & $\gradient{0.44}$ \color{lighter-color}\tiny{$\pm .05$} & $\gradient{0.91}$ \color{lighter-color}\tiny{$\pm .01$} & $\gradient{0.81}$ \color{lighter-color}\tiny{$\pm .01$} \\ 
    GPT-4o  & $\gradient{0.62}$ \color{lighter-color}\tiny{$\pm .00$} & $\gradient{0.87}$ \color{lighter-color}\tiny{$\pm .01$} & $\gradient{0.62}$ \color{lighter-color}\tiny{$\pm .07$} & $\gradient{0.82}$ \color{lighter-color}\tiny{$\pm .01$} & $\gradient{0.77}$ \color{lighter-color}\tiny{$\pm .03$} & $\gradient{0.88}$ \color{lighter-color}\tiny{$\pm .04$} & $\gradient{0.95}$ \color{lighter-color}\tiny{$\pm .01$} & $\gradient{0.84}$ \color{lighter-color}\tiny{$\pm .01$} \\
    Claude  & $\gradient{0.98}$ \color{lighter-color}\tiny{$\pm .00$} & $\gradient{0.92}$ \color{lighter-color}\tiny{$\pm .01$} & $\gradient{0.33}$ \color{lighter-color}\tiny{$\pm .08$} & $\gradient{0.59}$ \color{lighter-color}\tiny{$\pm .02$} & $\gradient{0.28}$ \color{lighter-color}\tiny{$\pm .03$} & $\gradient{0.36}$ \color{lighter-color}\tiny{$\pm .06$} & $\gradient{0.92}$ \color{lighter-color}\tiny{$\pm .01$} & $\gradient{0.84}$ \color{lighter-color}\tiny{$\pm .01$} \\
    Mixtral & $\gradient{0.98}$ \color{lighter-color}\tiny{$\pm .01$} & $\gradient{0.85}$ \color{lighter-color}\tiny{$\pm .01$} & $\gradient{0.24}$ \color{lighter-color}\tiny{$\pm .08$} & $\gradient{0.56}$ \color{lighter-color}\tiny{$\pm .03$} & $\gradient{0.49}$ \color{lighter-color}\tiny{$\pm .03$} & $\gradient{0.56}$ \color{lighter-color}\tiny{$\pm .04$} & $\gradient{0.86}$ \color{lighter-color}\tiny{$\pm .01$} & $\gradient{0.80}$ \color{lighter-color}\tiny{$\pm .00$} \\
    Llama3  & $\gradient{0.87}$ \color{lighter-color}\tiny{$\pm .01$} & $\gradient{0.49}$ \color{lighter-color}\tiny{$\pm .02$} & $\gradient{0.16}$ \color{lighter-color}\tiny{$\pm .07$} & $\gradient{0.37}$ \color{lighter-color}\tiny{$\pm .03$} & $\gradient{0.33}$ \color{lighter-color}\tiny{$\pm .03$} & $\gradient{0.33}$ \color{lighter-color}\tiny{$\pm .05$} & $\gradient{0.86}$ \color{lighter-color}\tiny{$\pm .02$} & $\gradient{0.79}$ \color{lighter-color}\tiny{$\pm .01$} \\\midrule
    & \multicolumn{2}{c}{{\hyperref[elegy]{Elegy}}
} & \multicolumn{2}{c}{{\hyperref[blank-verse]{Blank Verse}}
} & \multicolumn{2}{c}{{\hyperref[dramatic]{Dramatic Monologue}}
} & \multicolumn{2}{c}{{\hyperref[ekphra]{Ekphrastic}
}} \\\cmidrule(lr){2-3}\cmidrule(lr){4-5}\cmidrule(lr){6-7}\cmidrule(lr){8-9}
               model & Memorized & Not Memorized & Memorized & Not Memorized & Memorized & Not Memorized & Memorized & Not Memorized \\\midrule
    GPT-3.5 & $\gradient{0.66}$ \color{lighter-color}\tiny{$\pm .03$} & $\gradient{0.55}$ \color{lighter-color}\tiny{$\pm .02$} & \gradient{0.72} \color{lighter-color}\tiny{$\pm .05$}  & $\gradient{0.32}$ \color{lighter-color}\tiny{$\pm .05$} & $\gradient{0.65}$ \color{lighter-color}\tiny{$\pm .05$} & $\gradient{0.39}$ \color{lighter-color}\tiny{$\pm .05$} & $\gradient{0.36}$ \color{lighter-color}\tiny{$\pm .03$} & $\gradient{0.66}$ \color{lighter-color}\tiny{$\pm .03$} \\
    GPT-4   & $\gradient{0.71}$ \color{lighter-color}\tiny{$\pm .04$} & $\gradient{0.72}$ \color{lighter-color}\tiny{$\pm .02$} & $\gradient{0.75}$ \color{lighter-color}\tiny{$\pm .06$} & $\gradient{0.36}$ \color{lighter-color}\tiny{$\pm .06$} & $\gradient{0.77}$ \color{lighter-color}\tiny{$\pm .02$} & $\gradient{0.63}$\color{lighter-color}\tiny{$\pm .02$} & $\gradient{0.68}$ \color{lighter-color}\tiny{$\pm .04$} & $\gradient{0.71}$ \color{lighter-color}\tiny{$\pm .05$} \\
    GPT-4o  & $\gradient{0.78}$ \color{lighter-color}\tiny{$\pm .04$} & $\gradient{0.75}$ \color{lighter-color}\tiny{$\pm .03$} & $\gradient{0.90}$ \color{lighter-color}\tiny{$\pm .03$} & $\gradient{0.57}$ \color{lighter-color}\tiny{$\pm .03$} & $\gradient{0.78}$ \color{lighter-color}\tiny{$\pm .02$} & $\gradient{0.64}$ \color{lighter-color}\tiny{$\pm .03$} & $\gradient{0.63}$ \color{lighter-color}\tiny{$\pm .03$} & $\gradient{0.70}$ \color{lighter-color}\tiny{$\pm .03$} \\
    Claude  & $\gradient{0.72}$ \color{lighter-color}\tiny{$\pm .04$} & $\gradient{0.62}$ \color{lighter-color}\tiny{$\pm .03$} & $\gradient{0.77}$ \color{lighter-color}\tiny{$\pm .03$} & $\gradient{0.58}$ \color{lighter-color}\tiny{$\pm .03$} & $\gradient{0.63}$ \color{lighter-color}\tiny{$\pm .01$} & $\gradient{0.38}$ \color{lighter-color}\tiny{$\pm .01$} & $\gradient{0.68}$ \color{lighter-color}\tiny{$\pm .04$} & $\gradient{0.76}$ \color{lighter-color}\tiny{$\pm .04$} \\
    Mixtral & $\gradient{0.75}$ \color{lighter-color}\tiny{$\pm .03$} & $\gradient{0.65}$ \color{lighter-color}\tiny{$\pm .03$} & $\gradient{0.60}$ \color{lighter-color}\tiny{$\pm .04$} & $\gradient{0.30}$ \color{lighter-color}\tiny{$\pm .04$} & \gradient{0.61} \color{lighter-color}\tiny{$\pm .02$} & $\gradient{0.40}$ \color{lighter-color}\tiny{$\pm .02$} & $\gradient{0.54}$ \color{lighter-color}\tiny{$\pm .04$} & $\gradient{0.67}$ \color{lighter-color}\tiny{$\pm .04$} \\
    Llama3  & $\gradient{0.71}$ \color{lighter-color}\tiny{$\pm .04$} & $\gradient{0.65}$ \color{lighter-color}\tiny{$\pm .02$} & $\gradient{0.71}$ \color{lighter-color}\tiny{$\pm .04$} & $\gradient{0.35}$ \color{lighter-color}\tiny{$\pm .04$} & $\gradient{0.74}$ \color{lighter-color}\tiny{$\pm .02$} & $\gradient{0.40}$ \color{lighter-color}\tiny{$\pm .02$} & $\gradient{0.62}$ \color{lighter-color}\tiny{$\pm .03$} & $\gradient{0.67}$ \color{lighter-color}\tiny{$\pm .03$} \\\bottomrule
    \end{tabular}
    }
    \caption{
    LLM performance (F1 scores) by model for the eight most common forms, where the prompt only includes the text of the poem. Results are broken down by whether the poems were likely \textbf{memorized} or \textbf{not memorized} by GPT-4 (see Section \ref{memorization}). Standard deviations are shown for 20 bootstrapped samples of poems.}
        \label{table:memorization}
\end{table*}

\begin{table*}[t]
    \scriptsize
    \centering
    \begin{tabular}{lcccccccccccccccccc}
    \toprule
    & \multicolumn{3}{c}{{\hyperref[table:poetic_features]{Rhyme}}} & \multicolumn{3}{c}{{\hyperref[table:poetic_features]{Repetition}}} & \multicolumn{3}{c}{{\hyperref[table:poetic_features]{Meter}}} \\\cmidrule(lr){2-4}\cmidrule(lr){5-7}\cmidrule(lr){8-10}
               model & f1 & precision & recall & f1 & precision & recall & f1 & precision & recall \\\midrule
    Claude  & \gradient{0.77} & 0.83 & 0.75 & \gradient{0.52} & 0.51 & 0.55 & \gradient{0.79} & 0.88 & 0.74 \\
    GPT-3.5 & \gradient{0.83} & 0.82 & 0.90 & \gradient{0.39} & 0.59 & 0.45 & \gradient{0.77} & 0.82 & 0.75 \\
    GPT-4   & \gradient{0.92} & 0.93 & 0.91 & \gradient{0.82} & 0.77 & 0.91 & \gradient{0.82} & 0.87 & 0.81 \\
    GPT-4o  & \gradient{0.94} & 0.93 & 0.95 & \gradient{0.75} & 0.66 & 0.91 & \gradient{0.88} & 0.91 & 0.85 \\
    Llama3  & \gradient{0.64} & 0.61 & 0.85 & \gradient{0.31} & 0.40 & 0.28 & \gradient{0.67} & 0.71 & 0.66 \\
    Mixtral & \gradient{0.71} & 0.88 & 0.68 & \gradient{0.54} & 0.71 & 0.55 & \gradient{0.66} & 0.79 & 0.65 \\\midrule
    & \multicolumn{3}{c}{{\hyperref[table:poetic_features]{Fixed Topic}}} & \multicolumn{3}{c}{{\hyperref[table:poetic_features]{Fixed Length}}} & \multicolumn{3}{c}{{\hyperref[table:poetic_features]{Visual Form}}} \\\cmidrule(lr){2-4}\cmidrule(lr){5-7}\cmidrule(lr){8-10}
               model & f1 & precision & recall & f1 & precision & recall & f1 & precision & recall \\\midrule
    Claude  & \gradient{0.58} & 0.61 & 0.61 & \gradient{0.65} & 0.72 & 0.71 & \gradient{0.39} & 0.56 & 0.31 \\
    GPT-3.5 & \gradient{0.50} & 0.51 & 0.55 & \gradient{0.60} & 0.74 & 0.66 & \gradient{0.30} & 0.54 & 0.22 \\
    GPT-4   & \gradient{0.65} & 0.69 & 0.64 & \gradient{0.79} & 0.85 & 0.82 & \gradient{0.58} & 0.67 & 0.55 \\
    GPT-4o  & \gradient{0.65} & 0.68 & 0.65 & \gradient{0.82} & 0.82 & 0.87 & \gradient{0.64} & 0.65 & 0.62 \\
    Llama3  & \gradient{0.53} & 0.53 & 0.53 & \gradient{0.50} & 0.64 & 0.61 & \gradient{0.28} & 0.59 & 0.19 \\
    Mixtral & \gradient{0.57} & 0.60 & 0.60 & \gradient{0.60} & 0.80 & 0.60 & \gradient{0.34} & 0.51 & 0.28 \\
    \bottomrule
    \end{tabular}
    \caption{LLM performance by model for the \textbf{poetic features}, where the prompt includes only the text of the poem.}
    \label{table:features-results}
\end{table*}

We compare the performance of six diverse, state-of-the-art LLMs on the task of identifying more than 20 poetic forms and formal elements from a list of possible options. We test three iterations of the GPT models---\textbf{GPT-3.5 Turbo}, \textbf{GPT-4} \cite{openai_gpt-4_2024_etal}, and \textbf{GPT-4o} \cite{openai_hello_2024}.
We also test \textbf{Claude 3 Sonnet} \cite{anthropic_claude_2024}, \textbf{Llama 3} \cite{meta_llama3_2024}, and the open-source \textbf{Mixtral 8x22B} \cite{mistral_ai_mixtral_2024}.

We experiment with four different zero-shot prompt types, showing the model different amounts of the poem and/or contextual information. 
Our goal in this study was not to design a state-of-the-art form detection system via training or few-shot examples but rather to probe the models for their current capabilities.
We prompt the model with 1) only the text of the poem; 2) only the title and author; 3) only the first line; 4) only the last line. We use these different prompts to test for memorization and to better understand how different aspects of a poem may impact  performance.

We additionally ask the model to provide both an elaborated and one-word rationale for its choice as well as a confidence score. An example prompt and response is included in \ref{example-prompt}.

\subsection{Results of Form Classification by LLMs}

\begin{figure*}[t]
    \centering
    \includegraphics[width=\textwidth]{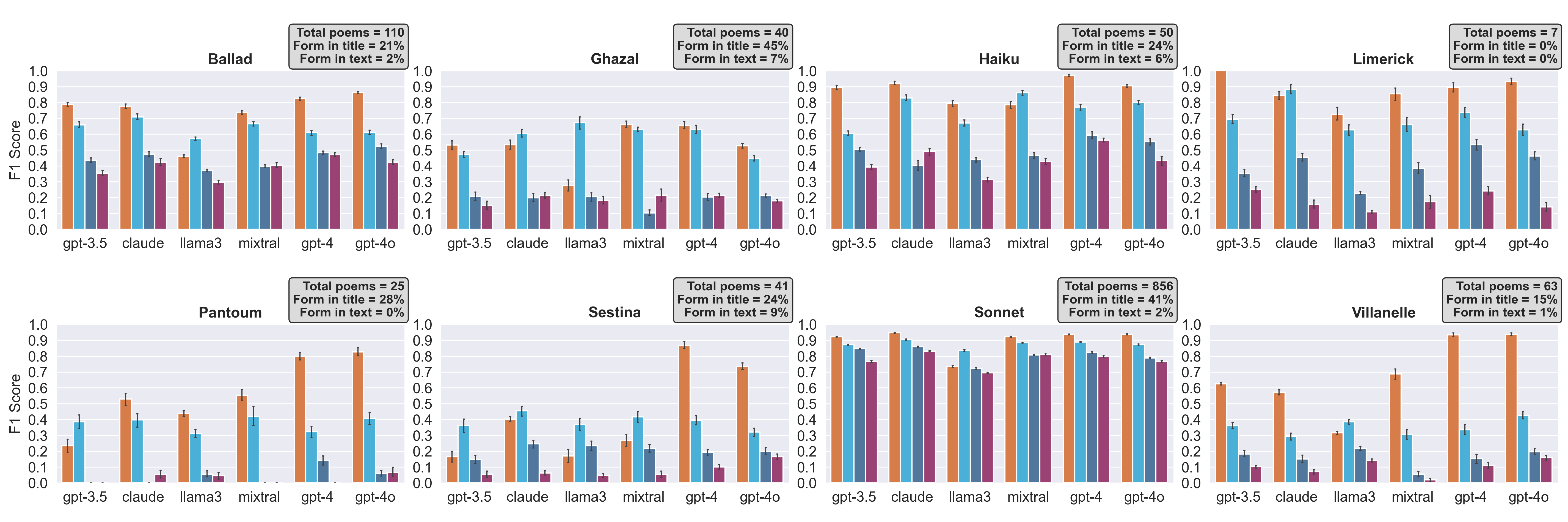}
    \includegraphics[width=0.12\textwidth]{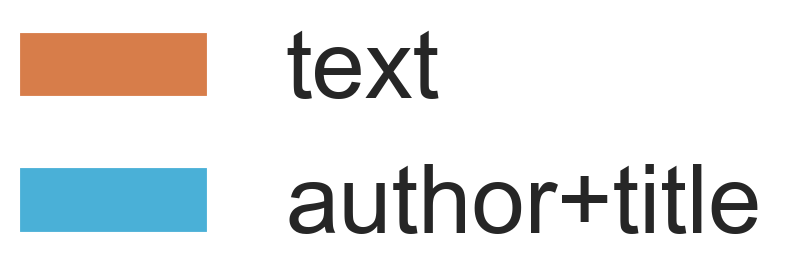}
    \includegraphics[width=0.12\textwidth]{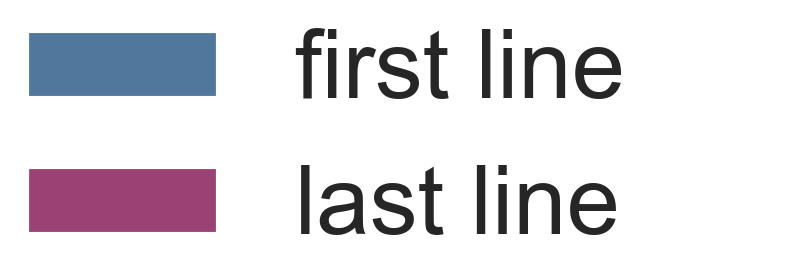}
    \caption{
    Poetic form classification results (F1 scores) for \textbf{fixed forms} when prompted with only the \textit{text of the poem}; only the \textit{author and title}; only the \textit{first line}; only the \textit{last line}.
    Error bars indicate standard deviation across 20 bootstrapped samples of poems. 
    See Figures \ref{fig:unfixed-form-performance-prompt}-\ref{fig:formal-elements-performance-prompt} in Appendix \ref{appendix-additional-form-results} for \textbf{unfixed forms} and \textbf{formal elements} results.}
    \label{fig:fixed-form-performance-prompt}
\end{figure*}

When prompted with only a poem's text, the LLMs perform better overall on the \textit{fixed} poetic forms than on the \textit{unfixed} forms or \textit{formal elements}. Performance for \textit{sonnets} and \textit{haiku} is particularly high, with F1 scores near or over $0.9$ for all models except Llama 3 (Table \ref{table:evaluation-results}). 
Averaging model performance by poetic feature (Table \ref{table:features-results}) suggests that the models may identify forms with rhyme, meter, and fixed length more easily (sonnets depend on all three, and haiku on length and syllable count). 

The models generally struggle to identify forms based on repetition, such as \textit{sestinas} (repeats same six endwords in specific pattern), \textit{villanelles} (first and third lines repeat in specific pattern), or \textit{pantoums} (second and fourth lines repeat in specific pattern).
But GPT-4 and GPT-4o do well in this rarer category, especially with \textit{sestinas} (F1=$0.87$; $0.73$), \textit{villanelles} (F1=$0.93$; $0.92$), and \textit{pantoums} (F1=$0.81$; $0.82$). This marks significant improvement from GPT-3.5 (F1=$0.17$, $0.62$, $0.20$) and is substantially stronger than Claude 3 Sonnet (F1=$0.41$, $0.58$, $0.53$), Mixtral 8x22B (F1=0.$26$, $0.69$, $0.56$), and Llama 3 (F1=$0.17$, $0.32$, $0.46$). 

Poetic forms based on topic prove more difficult for the models, depending on the topic (Table \ref{table:poetic_features}, \ref{table:unfixed-forms-performance}). Forms centered on more concrete subjects like death (\textit{elegy}) and art (\textit{ars poetica}, \textit{ekphrasis}) are more often recognized, while poems about abstract ideas and styles like \textit{aubades} and \textit{odes} are less so. 

There are fewer forms in our dataset that depend on visual features, but most models except GPT-4 and GPT-4o falter with them, namely with \textit{concrete or pattern poetry} (i.e. poems that rely on visual and typographical elements for their structure) and \textit{prose poetry} (i.e. poems that don't have line breaks and look like prose).

\subsection{Investigating Memorization Issues}

When prompted with only the author and title of a poem (and not the text), the models achieve nearly as high or higher classification performance in certain categories (see Figures \ref{fig:fixed-form-performance-prompt}, \ref{fig:unfixed-form-performance-prompt}). For sonnets, all the models achieve F1 scores of $0.85$ or higherwith only the title and author, and scores of $0.70$ or higher with only the first or last line. These results suggest possible memorization issues. However, more than 40\% of the sonnets also include the word ``sonnet'' in their title. The models indeed perform better with the author/title prompt on forms often named in their titles,  like \textit{aubade} (56\%) and \textit{ode} (48\%) (see Figure \ref{fig:unfixed-form-performance-prompt} in Appendix \ref{appendix-additional-form-results}).

We also compare model performance between poems from our dataset (major online poetry websites) and a smaller sample of manually digitized poems found only in print books (see Figure \ref{fig:digitzed} in Appendix \ref{appendix-additional-memorization-results}). We see both improvements and declines in accuracy across different forms, which seem largely dependent on the makeup of poems in the samples. For example, classification accuracy for \textit{sonnets} drops the most dramatically in our hand-digitized sample, but many of these sonnets come from a single author with a distinctive and unconventional style (thus posing greater difficulty).

The impact of explicit memorization is also unclear. When we compare performance between poems that are memorized and not memorized by GPT-4 (\S\ref{memorization}), we see significant decreases with unmemorized poems (e.g., \textit{blank verse}, \textit{dramatic monologue}) (see Table \ref{table:memorization}). This suggests that memorization of a poem may enhance its form classification. But with certain poetic forms, we also see improvements with unmemorized poems (e.g., \textit{prose poem}, \textit{couplet}), though these forms also have an uneven distribution of memorized vs. unmemorized poems (for \textit{prose poems}, 56 vs. 486).  
Analyzing classification performance based on presence in Dolma's popular pretraining datasets yields similarly mixed results (Table \ref{table:not-in-dolma-performance}). We believe more work is needed in this area.  

\section{Discussion} 
\label{discussion}

\subsection{Implications for NLP Researchers}

Poetry poses unique challenges to NLP systems. 
Our form detection task helps point to potential strengths and weaknesses in LLMs that can be explored in future work.  
For example, while \textit{prose poems} and \textit{concrete poetry} would likely be the easiest forms for humans (even non-experts) to identify because of obvious visual cues (lack of line breaks; text in a certain shape), the models generally struggle with them (Table \ref{table:unfixed-forms-performance}). This suggests potential difficulty with white space and visual dimensions of text.
Our study also shows that poetic features can pose challenges for models in different configurations and combinations. For example, the models do better with \textit{pantoums} than \textit{ghazals} even though both hinge on repetition (Table \ref{table:unfixed-forms-performance}). But \textit{pantoums} contain repetition of entire lines while \textit{ghazals} contain repetition of a single endword, perhaps suggesting that repetition of longer sequences enables easier classification. 
Lastly, our audit of pretraining data holds important lessons for NLP researchers who are designing evaluation benchmarks, e.g., memorization is an uneven issue that is difficult to quantify, heightening the importance of open resources for auditing pretraining data.

\subsection{Implications for Poetry Researchers, Readers, and Digitized Collections}

Automatic poetic form detection has the potential to improve discoverability of poems in digital libraries and archives. Poems were often published in periodicals, collections, and anthologies; when these sources are digitized, finding individual texts becomes difficult. 
Reliable detection of verse forms could aid in the identification of poetic texts within digitized collections. 
Additionally, LLM evaluations may offer literary insight into the legibility and durability of different poetic forms. For example, LLMs' successful classification of \textit{sonnets} may provide further evidence for the form's status as ``an exceptionally transnational poetic design... dispersed throughout more of the modern world than any other type of Western lyric'' \cite{maxwell_introduction_2004}. 
Finally, this research has implications for scholarship on the circulation and reception of poems online.  
Analyzing which lines appear in training data offers insight into where poems appear on the internet and how they travel online.

\section{Related Work}

\subsection{Poetry Generation and Analysis}

Machine-generated poetry has been a focal area in NLP for many decades and has received renewed interest in the era of LLMs \cite{manurung_towards_2000, goncalo_oliveira_survey_2017, van_de_cruys_automatic_2020, ormazabal_poelm_2022, lewis_syllable_2021, hu_poetrydiffusion_2023, melanie-becquet_exploring_2024, yu_charpoet_2024}.   
The computational analysis of poetry, including form and features like rhyme and meter, has a similarly long history 
\cite{petrick_understanding_1977, yokoyama_digital_nodate, genzel_poetic_2010, anttila_phonological_2015, heuser_quadrismegistusprosodic_2024, haider_metrical_2021, abdibayev_bpomp_2021, abdibayev_automating_2021, de_la_rosa_alberti_2023v2}. 

Recent NLP work has specifically addressed LLMs' capacity to understand poetry. \citet{mahbub-etal-2023-unveiling} develop a task and dataset, ``PoemSum'' (3,011 poem summary/poem text pairs collected from PoemAnalysis.com and various websites), to evaluate how well LLMs can summarize poetry. 
We build on this work by focusing on a more specific sub-task (poetic form detection), by curating a dataset of poems tagged by form (thus attending to internal differences between poems), and by selecting poems from well-respected poetry institutions.

\subsection{Literary Genre/Form Classification}

Automatically classifying literary texts by genre has been an active area of research in both NLP and digital humanities (DH). Many studies have focused on classifying fictional prose writing genres in novels \citep{underwood_life_2016v2, wilkens_genre_2016v2}, while other work has focused on distinguishing between kinds of poetry, such as Greek epic vs. drama \cite{gianitsos_stylometric_2019} and various styles of spoken free verse \cite{baumann_style_2018}.

In DH, genre classification has often been used to highlight ambiguity.
\citet{long_literary_2016_v2} find that features of English-language haiku are statistically distinct, yet they emphasize the importance of misclassifications for examining how ``broadly distributed haiku’s influence was.''  \citet{rhody_topic_2012} similarly suggests that computational analysis of poetry ``works, in part, because of its failures.''  These scholars largely use classification to explore the fuzziness, as opposed to the rigidity, of genres and poetic forms. 
We do not fully explore this angle in our work, which is important for future research.

\section{Conclusion}

Our work audits current poetic capacities and training data in leading LLMs. 
We contribute the poetry evaluation task and release to the research community a dataset of 1.4k+ annotated public domain poems with accompanying metadata about their prevalence in popular training datasets. 
We also join \citet{orr_ai_2024} and others in cautioning the benchmark/task as the be-all and end-all framework for NLP research.
Poetry is a good example of a human output that purposely troubles neat categorization.
We encourage more work that builds nuance and ambiguity into humanistic benchmarks such as this one, as well as work that places value beyond this orientation.
Further research is also needed to study LLM poetic capacities in languages beyond English and to evaluate impacts on human creators. 

\section{Limitations}\label{limitations}

In this study, we focus mostly on English-language poetry that was written and published in Europe and North America. Further, we only consider poems that were tagged by the Poetry Foundation, the Academy of American Poets, or editors of particular poetry collections (see \ref{manually-digitized-bib}), leaving out many other possible forms as well as poems that do not adhere neatly to forms.

Poetry Foundation and the Academy of American Poets do not have a comprehensive or representative (in terms of gender, race, culture, geography) collections of poems, nor do the print anthologies we digitized. Additionally, most of the poems in these collections are \textit{not} tagged by form, and it is not always clear why some poems have tags and others do not. For example, on the Poetry Foundation website, Etheridge Knight and Sonia Sanchez, two late 20th-century poets associated with the Black Arts Movement, both wrote haiku series that include the word ``haiku'' in their titles, but they are not tagged as haiku on Poetry Foundation.

While we select these resources because they are well-respected poetry institutions, we do not know how exactly these tags were applied to the poems, or who put them there. There may be cases where we or others disagree with the tags, but we keep all tags as found on the websites in order to represent the perspective of these institutions.

On these websites, and thus within our dataset, there is also an uneven distribution of poems in each form, reflecting biases related to race, class, language, and culture. For example, the \textit{ghazal} is a poetic form that originated in Arabic and is popular in the Middle East and South Asia; however, ghazals are less popular, and less likely to be curated, in English-language contexts. 
\textit{Limericks} are another popular and pervasive genre of poetry, yet they are often considered an unsophisticated genre or ``light verse'' form, and thus there are few of them in this particular dataset.

There are also limitations to conceiving of poetic form as a single-label classification task, as a set of independent categories that a poem can belong to or not. Poetry is often valued for ambiguity, experimentation, and interpretive potential, so fitting neatly into a category is not necessarily what one looks for in poetic analysis. Poets also often mix and merge forms. For example, Gwendolyn Brooks developed the ``\href{https://poets.org/poem/sonnet-ballad}{Sonnet-Ballad},'' and Roger Sedarat has created the ``Sonnet Ghazal'' \cite{sedarat_ghazal_2011}. Our approach does not account for these kinds of hybrid forms. Further, form only exists in relation to content. As foundational English literary scholars \citet{brooks_understanding_1960} wrote, ``the reader, unlike a robot, must be able to recognize the dramatic implications of the form.'' These implications only come through when form is considered as part of a broader composition with numerous intertwined elements.

\section{Ethical Considerations}

Many of the poems that we asked the models to identify are currently under copyright. The poems from Poetry Foundation and Academy of American Poets are freely available online, but this is due to the fact that these institutions pay for copyright and compensate poets for their work, which is crucial for reproduction of recent texts. In the dataset we share, we only include poems that are in the public domain and whose authors died before 1929. In the U.S., copyright extends for 95 years after the date of first publication, so works published before 1929 are in the public domain.   

In using LLMs to evaluate poetry, there is a risk of reinforcing dominant understandings of poetic form and prosody. As has been well documented, LLMs can reproduce existing biases related to gender, race, class, and cultural background \cite{bender_dangers_2021},  and there is significant existing bias in discourse surrounding poetic form. \citet{strand_making_2000} emphasize that ``Women were often underrepresented in poetry in the sixteenth, seventeenth, and eighteenth centuries'' and were ``absent—whether in retrospect or reality... from the festival of form that poetry became in those centuries.'' And \citet{shockley_renegade_2011} notes that the ``discourse around innovative and avant-garde poetry in the U.S.,'' which has often emphasized discussions of form, ``has historically constructed these categories as implicitly `white,''' pointing out that ``African American poets, even when they were involved in, perhaps central to, now-canonical avant-garde movements have been marginalized or erased from literary histories.'' 

These literary histories inform which works are included in anthologies and incorporated into digital collections, and they also influence training data. \citet{dsouza_chatbot_2023} have shown that inclusion in the 1983 edition of the \textit{Norton Anthology of Poetry} was the best predictor of poem memorization in ChatGPT. This anthology represents a traditional view of the English poetic canon, favoring historical works published in the U.K. and the U.S., and excluding important works by women authors, Black and Indigenous authors and authors of color, and authors working outside Europe and North America. If the performance of LLMs improves in relation to poetic form evaluation, whose versions of form will be reproduced?

Given the complex cultural, historical, and textual conditions from which poetic forms emerge, as well as the centuries-long discourse surrounding how to label, categorize, and analyze form, this work requires domain expertise, and domain experts should be included in discussions about benchmarks for complex creative and interpretive tasks. At the same time, domain experts may have hesitations about this kind of collaboration, given the widespread use of copyrighted material in training data, and the risks LLMs pose to authors, whose work is fundamental to literary studies.

\section{Acknowledgements}

We would like to thank Dany Haddad, Yanai Elazar, and Kyle Lo for their help with Dolma and \textsc{WIMBD}, as well as Elizabeth Gronski for her help with early versions of this work. We are also grateful to Greg Yauney, David Mimno, and Matthew Wilkens for formative conversations and feedback about this project. This work was supported by the NEH AI for Humanists project.

\bibliography{custom}

\clearpage
\newpage
\appendix
\onecolumn

\section{Appendix}
\label{sec:appendix}

\subsection{Poetic Features by Form}
\label{appendix-poetic-features}

\begin{table}[h!]
\centering
\begin{tabular}{@{}|p{3cm}|c|c|c|c|c|c|@{}}
\hline
\textbf{Fixed Forms} & \textbf{Rhyme} & \textbf{Repetition} & \textbf{Meter} & \textbf{Fixed Topic} & \textbf{Fixed Length} & \textbf{Visual Form} \\
\hline
Ballad & & & \checkmark & & & \\
\hline
Ghazal & & \checkmark & & & & \\
\hline
Haiku & & & & \checkmark & \checkmark & \\
\hline
Limerick & \checkmark & & & & \checkmark & \\
\hline
Pantoum & & \checkmark & & & & \\
\hline
Sestina & & \checkmark & & & \checkmark & \\
\hline
Sonnet & \checkmark & & \checkmark & & \checkmark & \\
\hline
Villanelle & \checkmark & \checkmark & & & \checkmark & \\
\hline
\textbf{Formal Elements} & \textbf{Rhyme} & \textbf{Repetition} & \textbf{Meter} & \textbf{Fixed Topic} & \textbf{Fixed Length} & \textbf{Visual Form} \\
\hline
Blank Verse & & & \checkmark & & & \\
\hline
Common Measure & \checkmark & & \checkmark & & \checkmark & \\
\hline
Couplet & & & & & \checkmark & \\
\hline
Free Verse & & & \checkmark & & & \\
\hline
Quatrain & & & & & \checkmark & \\
\hline
Tercet & & & & & \checkmark & \\
\hline
\textbf{Unfixed Forms} & \textbf{Rhyme} & \textbf{Repetition} & \textbf{Meter} & \textbf{Fixed Topic} & \textbf{Fixed Length} & \textbf{Visual Form} \\
\hline
Ars Poetica & & & & \checkmark & & \\
\hline
Aubade & & & & \checkmark & & \\
\hline
Concrete Poetry & & & & & & \checkmark \\
\hline
Dramatic\newline Monologue & & & & \checkmark & & \\
\hline
Ekphrasis & & & & \checkmark & & \\
\hline
Elegy & & & & \checkmark & & \\
\hline
Ode & & & & \checkmark & & \\
\hline
Pastoral & & & & \checkmark & & \\
\hline
Prose Poem & & & & & & \checkmark \\
\hline
\end{tabular}
\caption{Distribution of poetic features by form. {\hyperref[table:features-results]{$\hookleftarrow$}} }
\label{table:poetic_features}
\end{table}

\subsection{Additional Poetic Form Detection Results}
\label{appendix-additional-form-results}

\begin{table}[H]
    \scriptsize
    \centering
    \resizebox{1\linewidth}{!}{ %
    \begin{tabular}
    {lcccccccccccccccccc}
    \toprule
    & \multicolumn{3}{c}{{\hyperref[ars]{Ars Poetic}}} & \multicolumn{3}{c}{{\hyperref[aubade]{Aubade}}} & \multicolumn{3}{c}{{\hyperref[concrete]{Concrete or Pattern Poetry}}} & \multicolumn{3}{c}{{\hyperref[dramatic]{Dramatic Monologue}}
} & \multicolumn{3}{c}{{\hyperref[ekphra]{Ekphrasis}}
} \\\cmidrule(lr){2-4}\cmidrule(lr){5-7}\cmidrule(lr){8-10}\cmidrule(lr){11-13}\cmidrule(lr){14-16}
               model & f1 & precision & recall & f1 & precision & recall & f1 & precision & recall & f1 & precision & recall & f1 & precision & recall \\\midrule
    GPT-3.5  & \gradient{0.39} & 0.30 & 0.55 & \gradient{0.27} & 0.24 & 0.31 & \gradient{0.07} & 0.33 & 0.04 & \gradient{0.48} & 0.50 & 0.45 & \gradient{0.61} & 0.59 & 0.63 \\
    GPT-4    & \gradient{0.64} & 0.66 & 0.63 & \gradient{0.44} & 0.55 & 0.38 & \gradient{0.29} & 0.50 & 0.21 & \gradient{0.68} & 0.56 & 0.87 & \gradient{0.71} & 0.88 & 0.59 \\
    GPT-4o   & \gradient{0.59} & 0.54 & 0.66 & \gradient{0.48} & 0.47 & 0.50 & \gradient{0.47} & 0.48 & 0.46 & \gradient{0.69} & 0.60 & 0.81 & \gradient{0.69} & 0.64 & 0.74 \\
    Claude   & \gradient{0.44} & 0.53 & 0.37 & \gradient{0.47} & 0.41 & 0.56 & \gradient{0.21} & 0.22 & 0.21 & \gradient{0.45} & 0.30 & 0.91 & \gradient{0.75} & 0.80 & 0.70 \\
    Mixtral  & \gradient{0.43} & 0.69 & 0.31 & \gradient{0.51} & 0.41 & 0.69 & \gradient{0.15} & 0.14 & 0.17 & \gradient{0.47} & 0.34 & 0.77 & \gradient{0.65} & 0.79 & 0.55 \\
    Llama3   & \gradient{0.41} & 0.29 & 0.69 & \gradient{0.34} & 0.38 & 0.31 & \gradient{0.20} & 0.25 & 0.17 & \gradient{0.50} & 0.40 & 0.66 & \gradient{0.66} & 0.83 & 0.54 \\\midrule
    & \multicolumn{3}{c}{{\hyperref[elegy]{Elegy}}
} & \multicolumn{3}{c}{{\hyperref[ode]{Ode}}
} & \multicolumn{3}{c}{{\hyperref[pastoral]{Pastoral}}
} & \multicolumn{3}{c}{{\hyperref[prose]{Prose Poem}}
} \\\cmidrule(lr){2-4}\cmidrule(lr){5-7}\cmidrule(lr){8-10}\cmidrule(lr){11-13}
               model & f1 & precision & recall & f1 & precision & recall & f1 & precision & recall & f1 & precision & recall \\\midrule
    GPT-3.5  & \gradient{0.58} & 0.44 & 0.87 & \gradient{0.23} & 0.53 & 0.14 & \gradient{0.57} & 0.59 & 0.55 & \gradient{0.52} & 0.75 & 0.40 \\
    GPT-4    & \gradient{0.67} & 0.82 & 0.64 & \gradient{0.44} & 0.57 & 0.36 & \gradient{0.57} & 0.49 & 0.68 & \gradient{0.86} & 0.84 & 0.89 \\
    GPT-4o   & \gradient{0.76} & 0.77 & 0.75 & \gradient{0.43} & 0.75 & 0.30 & \gradient{0.68} & 0.75 & 0.61 & \gradient{0.80} & 0.82 & 0.78 \\
    Claude   & \gradient{0.65} & 0.81 & 0.55 & \gradient{0.45} & 0.44 & 0.46 & \gradient{0.49} & 0.58 & 0.43 & \gradient{0.56} & 0.89 & 0.41 \\
    Mixtral  & \gradient{0.68} & 0.62 & 0.75 & \gradient{0.46} & 0.50 & 0.43 & \gradient{0.58} & 0.52 & 0.65 & \gradient{0.53} & 0.89 & 0.41 \\
    Llama3   & \gradient{0.67} & 0.59 & 0.76 & \gradient{0.40} & 0.32 & 0.55 & \gradient{0.44} & 0.46 & 0.43 & \gradient{0.35} & 0.92 & 0.22 \\\bottomrule
    \end{tabular}
     }
    \caption{LLM performance by model for the \textbf{unfixed forms}, where the prompt includes only the poem text.}
    \label{table:unfixed-forms-performance}
\end{table}

\begin{table*}[th]
    \scriptsize
    \centering
    \begin{tabular}
    {lcccccccccccccccccc}
    \toprule
    & \multicolumn{3}{c}{{\hyperref[couplet]{Couplet}}
} & \multicolumn{3}{c}{{\hyperref[quatrain]{Quatrain}}
} & \multicolumn{3}{c}{{\hyperref[tercet]{Tercet}}
} \\\cmidrule(lr){2-4}\cmidrule(lr){5-7}\cmidrule(lr){8-10}
               model & f1 & precision & recall & f1 & precision & recall & f1 & precision & recall \\\midrule
    GPT-3.5  & \gradient{0.16} & 0.97 & 0.09 & \gradient{0.23} & 0.14 & 0.70 & \gradient{0.61} & 0.62 & 0.60 \\
    GPT-4    & \gradient{0.49} & 0.90 & 0.34 & \gradient{0.37} & 0.23 & 0.96 & \gradient{0.72} & 0.92 & 0.60 \\
    GPT-4o   & \gradient{0.72} & 0.96 & 0.58 & \gradient{0.48} & 0.33 & 0.90 & \gradient{0.81} & 0.84 & 0.78 \\
    Claude   & \gradient{0.30} & 1.00 & 0.18 & \gradient{0.31} & 0.19 & 0.83 & \gradient{0.81} & 0.73 & 0.90 \\
    Mixtral  & \gradient{0.51} & 0.99 & 0.35 & \gradient{0.27} & 0.18 & 0.56 & \gradient{0.69} & 0.55 & 0.90 \\
    Llama3   & \gradient{0.33} & 0.98 & 0.20 & \gradient{0.29} & 0.17 & 0.88 & \gradient{0.34} & 0.55 & 0.24 \\\midrule
    & \multicolumn{3}{c}{{\hyperref[blank-verse]{Blank Verse}}} & \multicolumn{3}{c}{{\hyperref[common-measure]{Common Measure}}} & \multicolumn{3}{c}{{\hyperref[free-verse]{Free Verse}}} \\\cmidrule(lr){2-4}\cmidrule(lr){5-7}\cmidrule(lr){8-10}
               model & f1 & precision & recall & f1 & precision & recall & f1 & precision & recall \\\midrule
    GPT-3.5  & \gradient{0.53} & 0.72 & 0.42 & \gradient{0.79} & 0.86 & 0.74 & \gradient{0.84} & 0.75 & 0.95 \\
    GPT-4    & \gradient{0.57} & 0.84 & 0.43 & \gradient{0.91} & 0.99 & 0.85 & \gradient{0.85} & 0.75 & 0.98 \\
    GPT-4o   & \gradient{0.73} & 0.92 & 0.61 & \gradient{0.96} & 0.95 & 0.97 & \gradient{0.89} & 0.83 & 0.96 \\
    Claude   & \gradient{0.68} & 0.70 & 0.67 & \gradient{0.65} & 1.00 & 0.48 & \gradient{0.88} & 0.81 & 0.95 \\
    Mixtral  & \gradient{0.46} & 0.54 & 0.41 & \gradient{0.36} & 1.00 & 0.22 & \gradient{0.83} & 0.72 & 0.97 \\
    Llama3   & \gradient{0.53} & 0.78 & 0.41 & \gradient{0.80} & 0.70 & 0.94 & \gradient{0.82} & 0.78 & 0.87 \\\bottomrule
    \end{tabular}
    \caption{LLM performance by model for the \textbf{formal elements}, where the prompt includes only the poem text.}
    \label{table:stanzas-meters-results}
\end{table*}

\begin{figure}[H]
    \centering
    \includegraphics[width=1\textwidth]{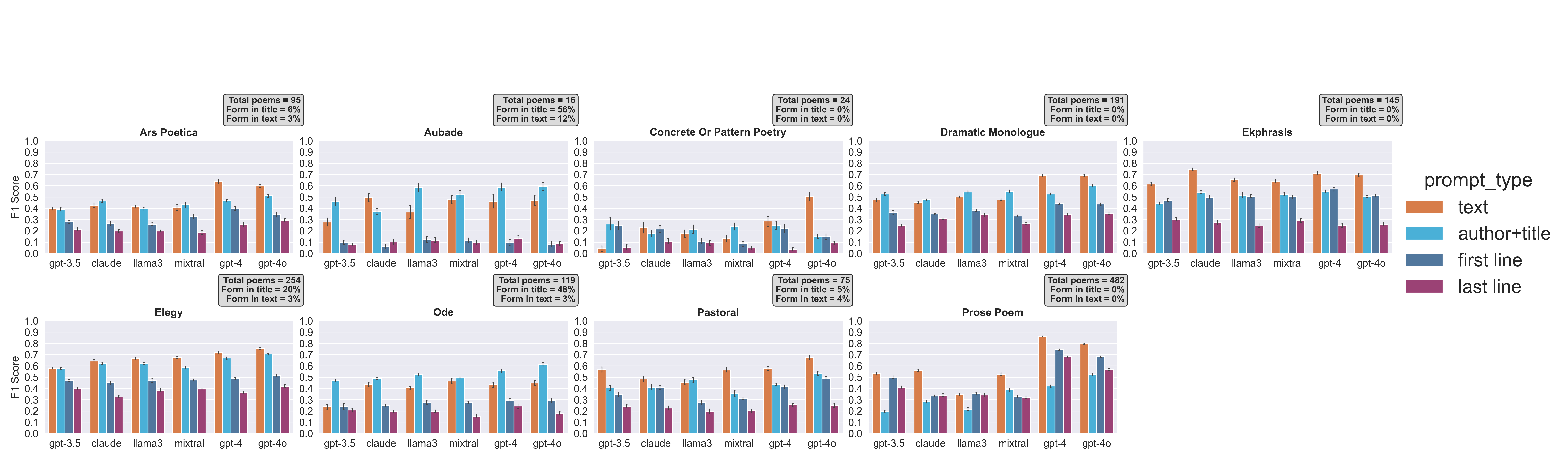}
    \caption{\textbf{Unfixed Forms — Poetry Foundation and Academy of American Poets.} These figures show LLM performance (F1 scores) by prompt type on the task of detecting poetic form (in the same way as the human annotation/institution it was collected from) by prompt type: with only the text of the poem; only the author and title; only the first line; only the last line. Error bars indicate standard deviation across 20 bootstrapped samples of poems.}
    \label{fig:unfixed-form-performance-prompt}
\end{figure}

\begin{figure}[H]
    \centering
    \includegraphics[width=1\textwidth]{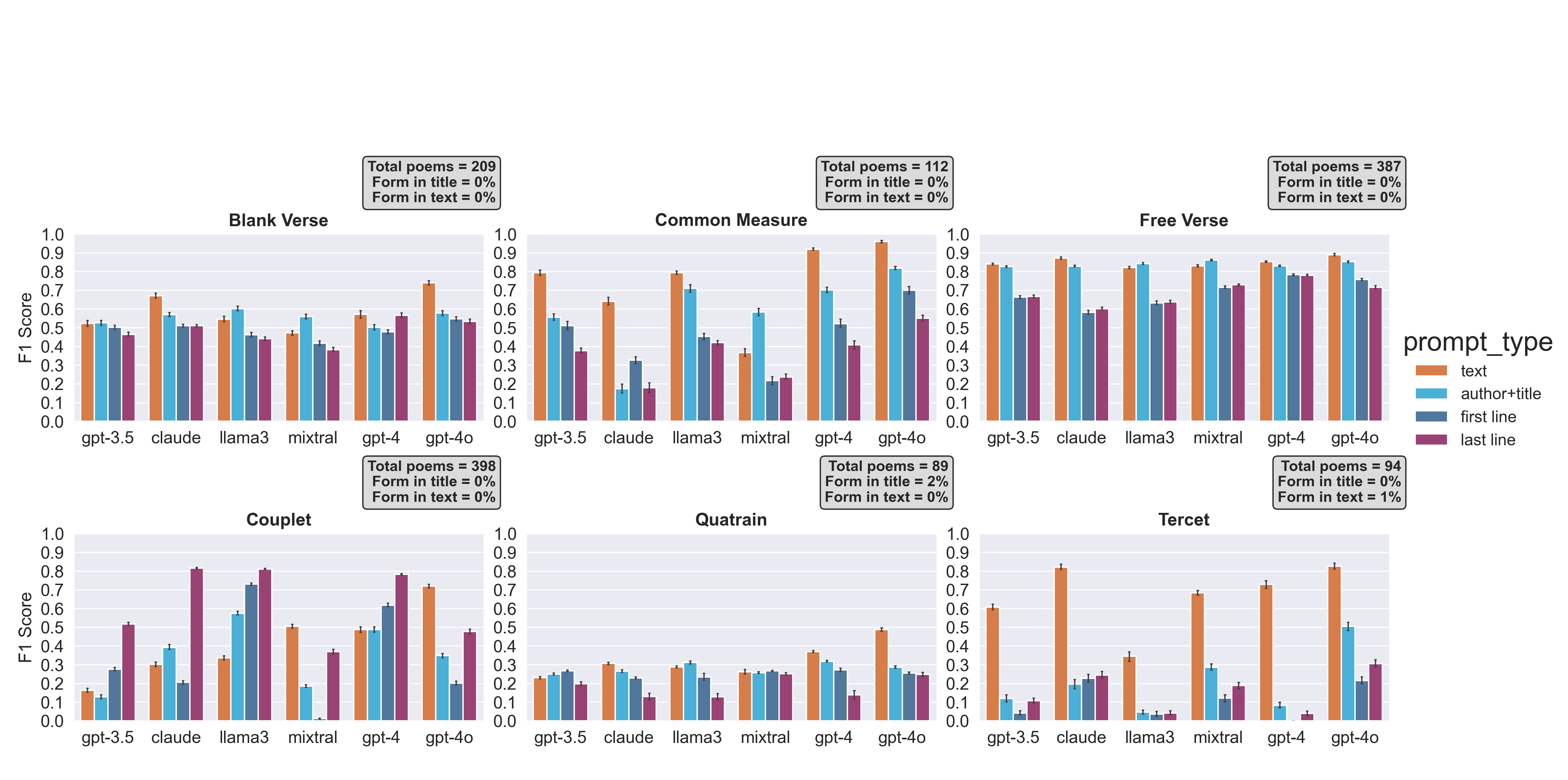}
    \caption{\textbf{Formal Elements — Poetry Foundation and Academy of American Poets.} These figures show LLM performance (F1 scores) by prompt type on the task of detecting a poem's form (in the same way as the human annotation/institution it was collected from) by prompt type: with only the text of the poem; only the author and title; only the first line; only the last line. Error bars indicate standard deviation across 20 bootstrapped samples of poems.}
    \label{fig:formal-elements-performance-prompt}
\end{figure}

\subsection{Additional Memorization Results}
\label{appendix-additional-memorization-results}

\begin{table}[H]
    \scriptsize
    \centering
    \begin{tabular}{lcccccccccccccccccccc}
    \toprule
    & \multicolumn{3}{c}{Sonnet} & \multicolumn{3}{c}{Limerick} & \multicolumn{3}{c}{Haiku} & \multicolumn{3}{c}{Ballad} \\\cmidrule(lr){2-4}\cmidrule(lr){5-7}\cmidrule(lr){8-10}\cmidrule(lr){11-13}
               model & f1 & precision & recall & f1 & precision & recall & f1 & precision & recall & f1 & precision & recall \\\midrule
    GPT-3.5 & \gradient{0.87} & 0.90 & 0.84 & \gradient{0.86} & 1.00 & 0.75 & \gradient{0.88} & 0.89 & 0.87 & \gradient{0.71} & 0.70 & 0.71 \\
    GPT-4   & \gradient{0.88} & 0.98 & 0.80 & \gradient{0.86} & 1.00 & 0.75 & \gradient{0.97} & 1.00 & 0.95 & \gradient{0.71} & 0.62 & 0.81 \\
    GPT-4o  & \gradient{0.87} & 0.99 & 0.78 & \gradient{1.00} & 1.00 & 1.00 & \gradient{0.88} & 0.94 & 0.82 & \gradient{0.77} & 0.82 & 0.72 \\
    Claude  & \gradient{0.92} & 0.93 & 0.91 & \gradient{0.75} & 1.00 & 0.60 & \gradient{0.92} & 1.00 & 0.85 & \gradient{0.63} & 0.88 & 0.49 \\
    Mixtral & \gradient{0.85} & 0.93 & 0.78 & \gradient{0.60} & 1.00 & 0.43 & \gradient{0.73} & 0.96 & 0.59 & \gradient{0.62} & 0.58 & 0.65 \\
    Llama3  & \gradient{0.49} & 0.99 & 0.33 & \gradient{0.60} & 1.00 & 0.43 & \gradient{0.73} & 0.96 & 0.59 & \gradient{0.30} & 0.20 & 0.65 \\\midrule
    & \multicolumn{3}{c}{Sestina} & \multicolumn{3}{c}{Villanelle} & \multicolumn{3}{c}{Pantoum} & \multicolumn{3}{c}{Ghazal} \\\cmidrule(lr){2-4}\cmidrule(lr){5-7}\cmidrule(lr){8-10}\cmidrule(lr){11-13}
               model & f1 & precision & recall & f1 & precision & recall & f1 & precision & recall & f1 & precision & recall \\\midrule
    GPT-3.5 & \gradient{0.20} & 0.80 & 0.11 & \gradient{0.60} & 0.44 & 0.94 & \gradient{0.22} & 0.60 & 0.14 & \gradient{0.56} & 0.49 & 0.66 \\
    GPT-4   & \gradient{0.86} & 0.84 & 0.89 & \gradient{0.95} & 1.00 & 0.90 & \gradient{0.82} & 0.72 & 0.95 & \gradient{0.64} & 0.49 & 0.91 \\
    GPT-4o  & \gradient{0.71} & 0.61 & 0.86 & \gradient{0.92} & 0.90 & 0.94 & \gradient{0.83} & 0.77 & 0.91 & \gradient{0.49} & 0.34 & 0.89 \\
    Claude  & \gradient{0.41} & 0.31 & 0.61 & \gradient{0.55} & 0.58 & 0.52 & \gradient{0.57} & 0.54 & 0.59 & \gradient{0.52} & 0.57 & 0.49 \\
    Mixtral & \gradient{0.29} & 1.00 & 0.17 & \gradient{0.72} & 0.79 & 0.66 & \gradient{0.55} & 0.48 & 0.64 & \gradient{0.68} & 0.60 & 0.77 \\
    Llama3  & \gradient{0.15} & 0.60 & 0.08 & \gradient{0.31} & 0.19 & 0.86 & \gradient{0.44} & 0.31 & 0.73 & \gradient{0.30} & 0.42 & 0.23 \\\bottomrule
    \end{tabular}
    \caption{\textbf{For poems likely not memorized by GPT-4.} LLM performance by model for the \textbf{fixed forms}, where the prompt includes only the poem text.}
    \label{table:not-memorized-performance}
\end{table}

\begin{figure}[H]
    \centering
\includegraphics[width=\textwidth, trim={0 0 0 4cm},clip]{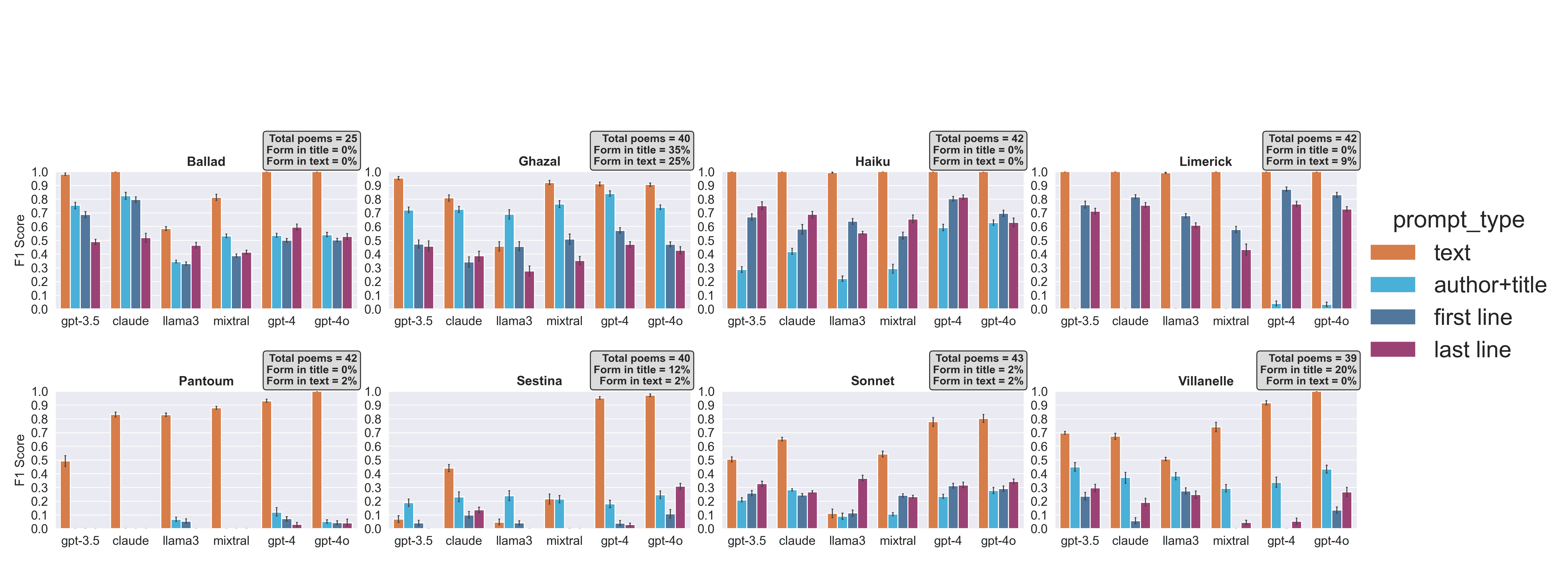}
    \caption{\textbf{Fixed Forms — Manually Digitized Poetry Books.} These figures show LLM performance (F1 scores) on the task of detecting a poem's form (in the same way as the human annotation/institution it was collected from) by prompt type: with only the text of the poem; only the author and title; only the first line; only the last line. Error bars indicate standard deviation across 20 bootstrapped samples of poems. The poems tested in this experiment were included in print books with little to no digital presence and manually digitized/transcribed by our team.}
    \label{fig:digitzed}
\end{figure}

\begin{table}[H]
    \scriptsize
    \centering
    \begin{tabular}{lcccccccccccccccccccc}
    \toprule
    & \multicolumn{3}{c}{Sonnet} & \multicolumn{3}{c}{Limerick} & \multicolumn{3}{c}{Haiku} & \multicolumn{3}{c}{Ballad} \\\cmidrule(lr){2-4}\cmidrule(lr){5-7}\cmidrule(lr){8-10}\cmidrule(lr){11-13}
               model & f1 & precision & recall & f1 & precision & recall & f1 & precision & recall & f1 & precision & recall \\\midrule
    GPT-3.5 & \gradient{0.91} & 0.92 & 0.91 & \gradient{1.00} & 1.00 & 1.00 & \gradient{0.89} & 0.90 & 0.88 & \gradient{0.78} & 0.82 & 0.75 \\
    GPT-4   & \gradient{0.94} & 0.98 & 0.90 & \gradient{0.89} & 0.80 & 1.00 & \gradient{1.00} & 1.00 & 1.00 & \gradient{0.87} & 0.85 & 0.89 \\
    GPT-4o  & \gradient{0.94} & 0.99 & 0.89 & \gradient{1.00} & 1.00 & 1.00 & \gradient{0.90} & 0.96 & 0.84 & \gradient{0.90} & 0.93 & 0.88 \\
    Claude  & \gradient{0.94} & 0.94 & 0.95 & \gradient{0.80} & 0.67 & 1.00 & \gradient{0.93} & 1.00 & 0.88 & \gradient{0.79} & 0.98 & 0.67 \\
    Mixtral & \gradient{0.92} & 0.95 & 0.88 & \gradient{0.89} & 0.80 & 1.00 & \gradient{0.84} & 0.96 & 0.75 & \gradient{0.76} & 0.73 & 0.79 \\
    Llama3  & \gradient{0.74} & 1.00 & 0.58 & \gradient{0.67} & 0.50 & 1.00 & \gradient{0.84} & 0.96 & 0.75 & \gradient{0.49} & 0.35 & 0.82 \\\midrule
    & \multicolumn{3}{c}{Sestina} & \multicolumn{3}{c}{Villanelle} & \multicolumn{3}{c}{Pantoum} & \multicolumn{3}{c}{Ghazal} \\\cmidrule(lr){2-4}\cmidrule(lr){5-7}\cmidrule(lr){8-10}\cmidrule(lr){11-13}
               model & f1 & precision & recall & f1 & precision & recall & f1 & precision & recall & f1 & precision & recall \\\midrule
    GPT-3.5 & \gradient{0.08} & 1.00 & 0.04 & \gradient{0.65} & 0.49 & 0.94 & \gradient{0.33} & 0.75 & 0.21 & \gradient{0.47} & 0.40 & 0.57 \\
    GPT-4   & \gradient{0.84} & 0.84 & 0.84 & \gradient{0.91} & 0.97 & 0.85 & \gradient{0.80} & 0.67 & 1.00 & \gradient{0.68} & 0.53 & 0.95 \\
    GPT-4o  & \gradient{0.77} & 0.69 & 0.88 & \gradient{0.93} & 0.89 & 0.97 & \gradient{0.82} & 0.70 & 1.00 & \gradient{0.50} & 0.35 & 0.90 \\
    Claude  & \gradient{0.38} & 0.30 & 0.52 & \gradient{0.56} & 0.54 & 0.59 & \gradient{0.59} & 0.62 & 0.57 & \gradient{0.49} & 0.50 & 0.48 \\
    Mixtral & \gradient{0.21} & 1.00 & 0.12 & \gradient{0.79} & 0.96 & 0.68 & \gradient{0.67} & 0.55 & 0.86 & \gradient{0.64} & 0.58 & 0.71 \\
    Llama3  & \gradient{0.14} & 0.67 & 0.08 & \gradient{0.31} & 0.19 & 0.88 & \gradient{0.59} & 0.44 & 0.86 & \gradient{0.25} & 0.36 & 0.19 \\\bottomrule
    \end{tabular}
    \caption{\textbf{For poems not found in Dolma.} LLM performance by model for the \textbf{formal elements}, where the prompt includes only the poem text.}
    \label{table:not-in-dolma-performance}
\end{table}

\subsection{Formative Study Results}
\label{formative-study}

Our formative study includes the following four poems.
\begin{itemize}
\itemsep-.1em
    \small{\item \textbf{Poem 1}: John Crowe Ransom's \href{https://www.poetryfoundation.org/poems/49146/piazza-piece}{``Piazza Piece''} (\textit{sonnet})
    \item \textbf{Poem 2}: Robert Browning's \href{https://www.poetryfoundation.org/poems/43773/prospice}{``Prospice''} (\textit{ballad}) 
    \item \textbf{Poem 3}: Natalie Diaz's \href{https://www.poetryfoundation.org/poems/56355/my-brother-at-3-am}{``My Brother at 3 A.M.''} (\textit{pantoum})
    \item \textbf{Poem 4}: Matthew Rohrer's \href{https://www.poetryfoundation.org/poetrymagazine/poems/57528/poem-written-with-buson-in-a-minute}{``Poem Written with Buson [`In a minute']''} (\textit{haiku)}.}
\end{itemize}

\begin{figure}[H]
    \centering
    \includegraphics[width=1\linewidth]{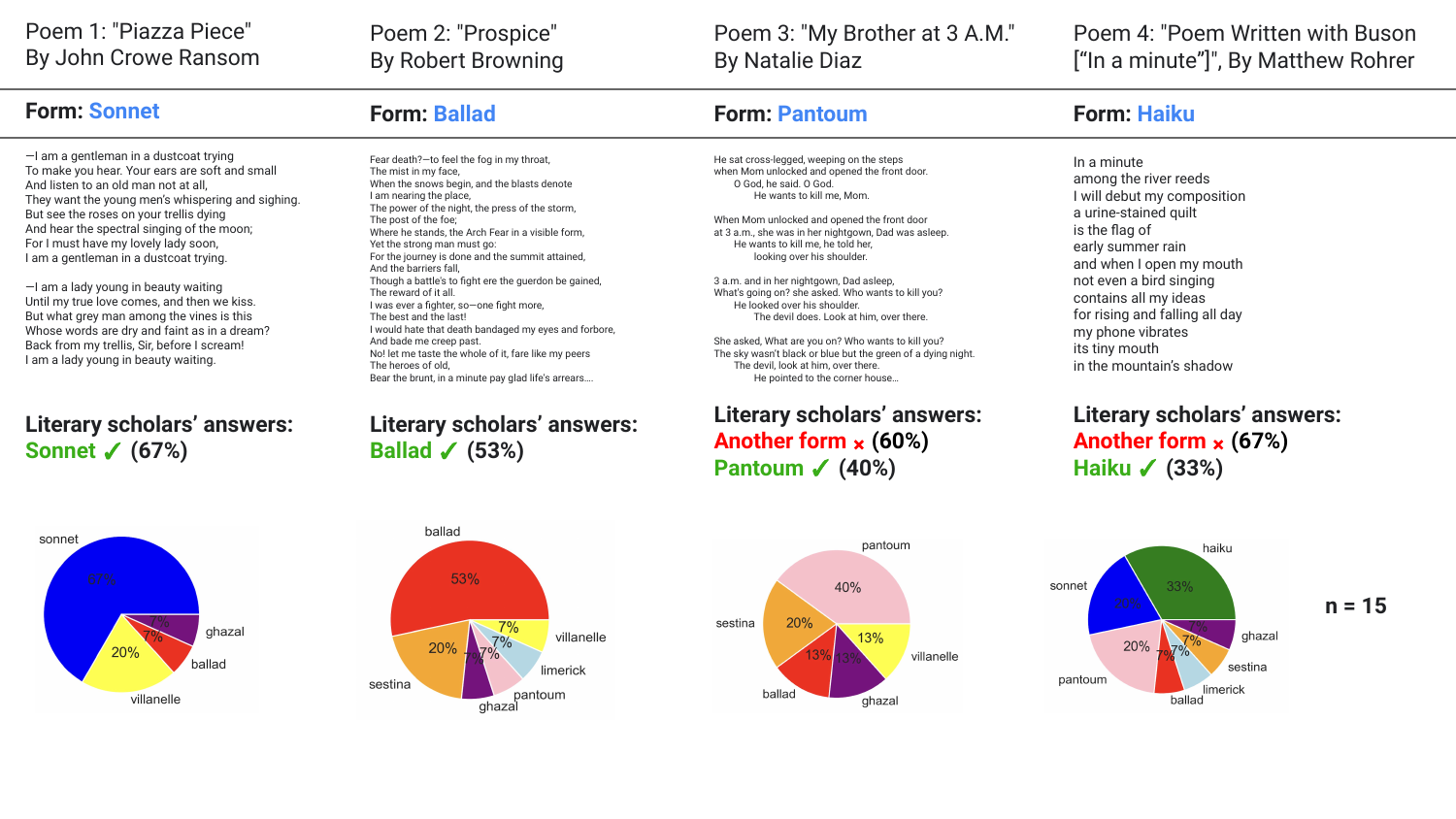}
    \caption{Results from formative study with human experts.}
    \label{fig:scholar-study}
\end{figure}

\subsection{Example Prompt and LLM Response} \label{example-prompt}

\textbf{Example Prompt:}
\begin{quote}

\textit{Read the following poem and then respond with the form of the poem based on the following verse forms: ['sonnet', 'ballad', 'villanelle', 'haiku', 'ghazal', 'sestina', 'pantoum', 'limerick'].\\
All of the poems have been tagged by experts as one of these forms. You must pick one of these options.}

\textit{Please also provide an elaborated rationale for why you think the poem is in this form, a one-word summary rationale, and a score ranking your confidence in your answer from 0 to 1.
Please report the single poetic form, elaborated rationale, and one-word rationale in the following format.}

\textit{1. Poetic Form: Ballad\\
2. Elaborated Rationale: This poem is a ballad because...\\
3. One-Word Summary: Meter\\
4. Confidence Score : 0.91}

\textit{1. Poetic Form: Sonnet\\
2. Elaborated Rationale: This poem is a sonnet because...\\
3. One-Word Summary: Meter\\
4. Confidence Score : 0.73\\}

\textit{Poem Text (in full): There was an old man of Thermopylæ,\\
Who never did anything properly;\\
But they said, ``If you choose, To boil eggs in your shoes,\\
You shall never remain in Thermopylæ.''
}
\newline
\newline
\textit{Pick ONE of these possible verse forms: ['sonnet', 'ballad', 'villanelle', 'haiku', 'ghazal', 'sestina', 'pantoum', 'limerick'].
}

\end{quote}
\textbf{Example Response (GPT-4):}
\begin{quote}
\textit{1. Poetic Form: Limerick\\
2. Elaborated Rationale: This poem is a limerick because it strictly follows the standard limerick rhyme scheme of AABBA. The poem begins with a character and a location (an old man of Thermopylæ), and it proceeds in a humorous, light-hearted manner typical of limericks. The first two lines are longer, the next two lines are shorter, and the final line is longer again, also within the parameters of a limerick.\\
3. One-Word Summary: Rhyme\\
4. Confidence Score : 0.98\\}
\end{quote}

\subsection{Manually Digitized Print Poetry Sources}\label{manually-digitized-bib}

\begin{raggedright}
John Barrell and John Bull. 1975. \textit{A Book of English pastoral verse}. Oxford University Press, New York.\\

Wilmon Brewer. 1937. \textit{Sonnets and Sestinas}. Cornhill Publishing Co., Boston.\\

Fred Cogswell. 1986. \textit{Meditations: 50 Sestinas. Ragweed Press}.\\

Clark Coolidge. 2012. \textit{88 Sonnets.} Fence Books, Albany, NY.\\

Dennis Daly. 2018. \textit{Pantoums.} Dos Madres Press, Loveland, Ohio.\\

Annie Finch, Marie-Elizabeth Mali, Annie Finch, and Julie Kane. 2012. \textit{Villanelles}. Alfred A. Knopf, New York.\\

Sandra M. Gilbert. 2001. \textit{Inventions of farewell: a book of elegies.} W.W. Norton \& Co., New York.\\

Robert Graves. 1957. \textit{English and Scottish ballads.} Heinemann, London.\\

Scott Gutterman, editor. 2015. \textit{Sunlight on the river: poems about paintings, paintings about poems.} Prestel, Munich.\\

Jim Kacian, Philip Rowland, and Allan Burns. 2013. \textit{Haiku in English: the first hundred years.} W.W. Norton \& Company, New York.\\

G. Legman. 1969. \textit{The Limerick: 1700 examples, with notes, variants, and index.} Bell Publishing Co., New York.\\

G. Legman. 1977. \textit{The New Limerick: 2750 Unpublished Examples, American and British.} Crown Publishers.\\ 

Bob Raczka. 2016. \textit{Wet cement: a mix of concrete poems.} Roaring Brook Press, New York.\\

Cor Van den Heuvel. 1986. \textit{The haiku anthology: haiku and senryu in English.} Simon \& Schuster, New York.\\

Joseph Warton. 1977.\textit{ Odes on various subjects (1746).} Scholars’ Facsimiles \& Reprints, Delmar, N.Y.\\

Eugene Wildman. 1967. \textit{The Chicago review anthology of concretism.} Swallow Press, Chicago.\\

Emmett Williams and Something Else Press. 1967.\textit{ An anthology of concrete poetry.} Something Else Press, New York.\\

Seishi Yamaguchi and Sono Uchida. 1993. \textit{The essence of modern haiku: 300 poems.} Mangajin, Inc., Atlanta, Georgia.\\

Kevin Young. 2010.\textit{ The art of losing: poems of grief and healing. }Bloomsbury USA, New York.\\

Thomas Perrin Harrison. 1968. \textit{The pastoral elegy: an anthology}. Octagon Books.\\
\end{raggedright}

\subsection{Memorization Prompt} 
\label{appendix-memorization-prompt}

\begin{quotation}
    \textit{What are the next five lines of the poem "<POEM\_TITLE>" by <AUTHOR\_NAME>?}

  \textit{  First Line: <FIRST\_LINE>}
    
    \textit{Next Lines:}
\end{quotation}

\clearpage

\subsection{Baseline Comparisons}
\label{appendix-baselines}

\begin{table*}[ht]
    \centering
    \footnotesize
    \begin{tabular}{@{}p{3.5cm}|p{2cm}@{\hspace{.75em}}p{2cm}@{\hspace{.75em}}p{2cm}@{}}
    \toprule
    \textbf{Form} & \textbf{Precision} & \textbf{Recall} & \textbf{F1}\\
    \midrule
    \textbf{Prose Poem}   
    & $0.79 \color{light-color}\pm 0.06$
    & $0.84 \color{light-color}\pm 0.08$
    & $0.81 \color{light-color}\pm 0.06$
    \\[1ex]
    \textbf{Sonnet}
    & $0.55 \color{light-color}\pm 0.12$
    & $0.66 \color{light-color}\pm 0.11$
    & $0.58 \color{light-color}\pm 0.07$
    \\[1ex]
    \textbf{Ekphrasis}
    & $0.34 \color{light-color}\pm 0.03$
    & $0.51 \color{light-color}\pm 0.07$
    & $0.40 \color{light-color}\pm 0.03$
    \\[1ex]
    \textbf{Ballad}   
    & $0.33 \color{light-color}\pm 0.10$
    & $0.53 \color{light-color}\pm 0.15$
    & $0.39 \color{light-color}\pm 0.10$
    \\[1ex]
    \textbf{Blank Verse}
    & $0.34 \color{light-color}\pm 0.16$
    & $0.34 \color{light-color}\pm 0.12$
    & $0.33 \color{light-color}\pm 0.12$
    \\[1ex]
    \textbf{Couplet}
    & $0.30 \color{light-color}\pm 0.04$
    & $0.38 \color{light-color}\pm 0.14$
    & $0.32 \color{light-color}\pm 0.07$
    \\[1ex]
    \textbf{Common Measure}
    & $0.40 \color{light-color}\pm 0.24$
    & $0.30 \color{light-color}\pm 0.23$
    & $0.31 \color{light-color}\pm 0.17$    
    \\[1ex]
    \textbf{Free Verse}
    & $0.27 \color{light-color}\pm 0.08$
    & $0.36 \color{light-color}\pm 0.13$
    & $0.29 \color{light-color}\pm 0.08$
    \\[1ex]
    \textbf{Dramatic Monologue}
    & $0.26 \color{light-color}\pm 0.15$
    & $0.32 \color{light-color}\pm 0.17$
    & $0.28 \color{light-color}\pm 0.15$
    \\[1ex]
    \textbf{Elegy}
    & $0.13 \color{light-color}\pm 0.11$
    & $0.04 \color{light-color}\pm 0.04$
    & $0.06 \color{light-color}\pm 0.05$
    \\[1ex]
    \textbf{Ode}
    & $0.17 \color{light-color}\pm 0.21$
    & $0.03 \color{light-color}\pm 0.04$
    & $0.05 \color{light-color}\pm 0.06$
    \\[1ex]
    \bottomrule
    \end{tabular}
    \caption{Cross-validation results ($k=5$) on the most common forms for a finetuned RoBERTa model.  We show the mean and standard deviation across the folds, and forms are ranked by their mean F1 score.
    } 
    \label{table:baseline-performance}
\end{table*}

\begin{table*}[ht]
    \centering
    \footnotesize
    \begin{tabular}{@{}p{3.5cm}|p{2cm}@{\hspace{.75em}}p{2cm}@{\hspace{.75em}}p{2cm}@{}}
    \toprule
    \textbf{Form} & \textbf{Precision} & \textbf{Recall} & \textbf{F1}\\
    \midrule
    \textbf{Sonnet}
    & $0.37 \color{light-color}\pm 0.10$
    & $0.50 \color{light-color}\pm 0.03$
    & $0.42 \color{light-color}\pm 0.06$
    \\[1ex]
    \textbf{Ballad}   
    & $0.31 \color{light-color}\pm 0.08$
    & $0.35 \color{light-color}\pm 0.14$
    & $0.32 \color{light-color}\pm 0.10$
    \\[1ex]
    \textbf{Blank Verse}
    & $0.32 \color{light-color}\pm 0.20$
    & $0.33 \color{light-color}\pm 0.10$
    & $0.28 \color{light-color}\pm 0.07$
    \\[1ex]
    \textbf{Dramatic Monologue}
    & $0.27 \color{light-color}\pm 0.15$
    & $0.29 \color{light-color}\pm 0.10$
    & $0.27 \color{light-color}\pm 0.12$
    \\[1ex]
    \textbf{Ekphrasis}
    & $0.29 \color{light-color}\pm 0.12$
    & $0.29 \color{light-color}\pm 0.09$
    & $0.26 \color{light-color}\pm 0.08$
    \\[1ex]
    \textbf{Free Verse}
    & $0.23 \color{light-color}\pm 0.08$
    & $0.22 \color{light-color}\pm 0.10$
    & $0.22 \color{light-color}\pm 0.09$
    \\[1ex]
    \textbf{Common Measure}
    & $0.19 \color{light-color}\pm 0.07$
    & $0.26 \color{light-color}\pm 0.11$
    & $0.21 \color{light-color}\pm 0.07$
    \\[1ex]
    \textbf{Prose Poem}   
    & $0.20 \color{light-color}\pm 0.02$
    & $0.22 \color{light-color}\pm 0.10$
    & $0.20 \color{light-color}\pm 0.05$
    \\[1ex]
    \textbf{Couplet}
    & $0.27 \color{light-color}\pm 0.13$
    & $0.18 \color{light-color}\pm 0.10$
    & $0.19 \color{light-color}\pm 0.08$
    \\[1ex]
    \textbf{Ode}
    & $0.22 \color{light-color}\pm 0.10$
    & $0.12 \color{light-color}\pm 0.03$
    & $0.15 \color{light-color}\pm 0.05$
    \\[1ex]
    \textbf{Elegy}
    & $0.20 \color{light-color}\pm 0.13$
    & $0.09 \color{light-color}\pm 0.06$
    & $0.11 \color{light-color}\pm 0.06$
    \\[1ex]
    \bottomrule
    \end{tabular}
    \caption{Cross-validation results ($k=5$) on the most common forms for an SVMs classifier with TF-IDF weighted unigram features. We show the mean and standard deviation across the folds, and forms are ranked by their mean F1 score.
    } 
    \label{table:baseline-performance-tfidf}
\end{table*}

\subsection{Misclassifications}

\begin{figure}[H]
    \centering
    \includegraphics[width=1\linewidth]{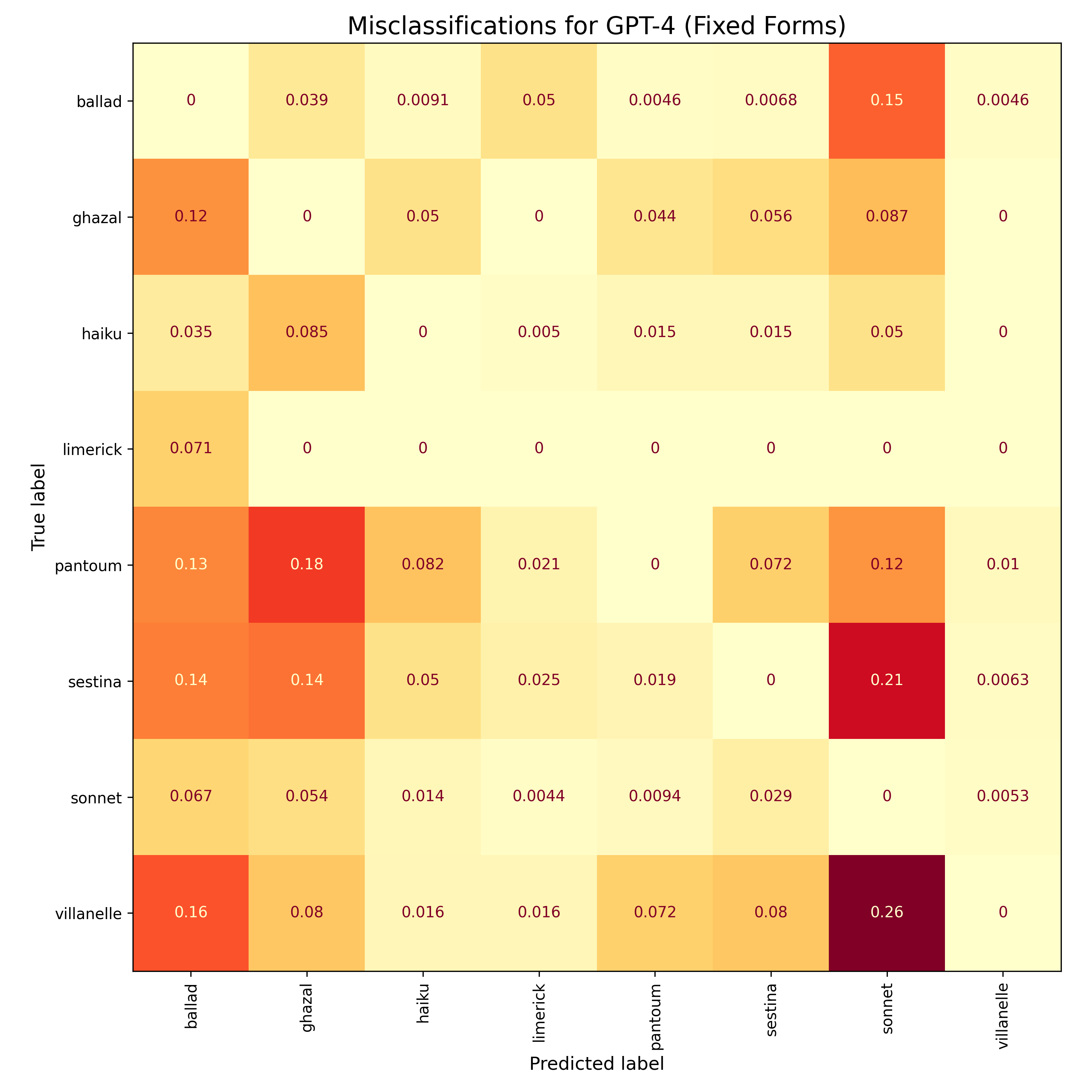}
    \caption{Misclassification proportions for fixed forms by GPT-4.}
    \label{fig:mis-fixed-forms1}
\end{figure}

\begin{figure}[H]
    \centering
    \includegraphics[width=1\linewidth]{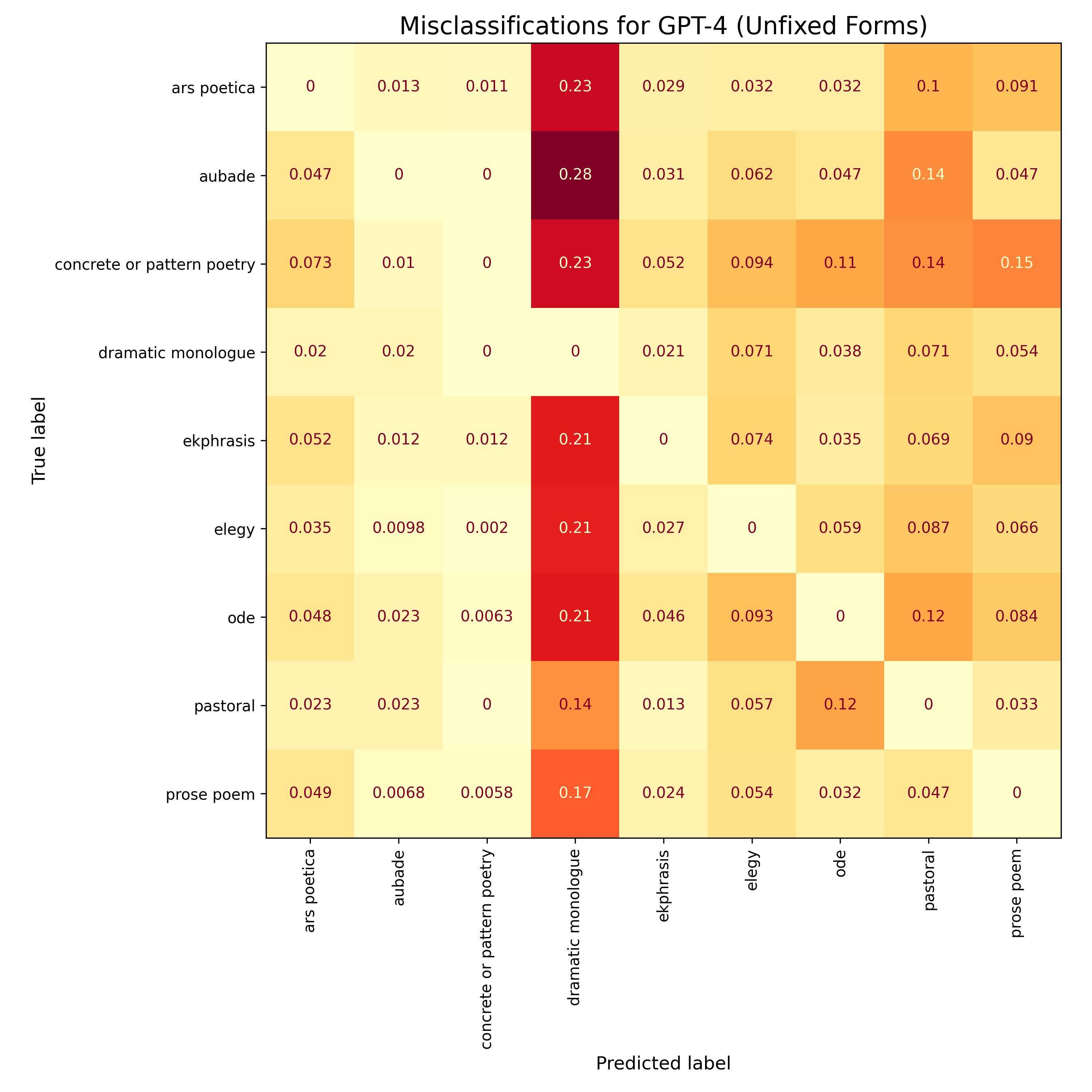}
    \caption{Misclassification proportions for unfixed forms by GPT-4.}
    \label{fig:mis-fixed-forms2}
\end{figure}

\begin{figure}[H]
    \centering
    \includegraphics[width=1\linewidth]{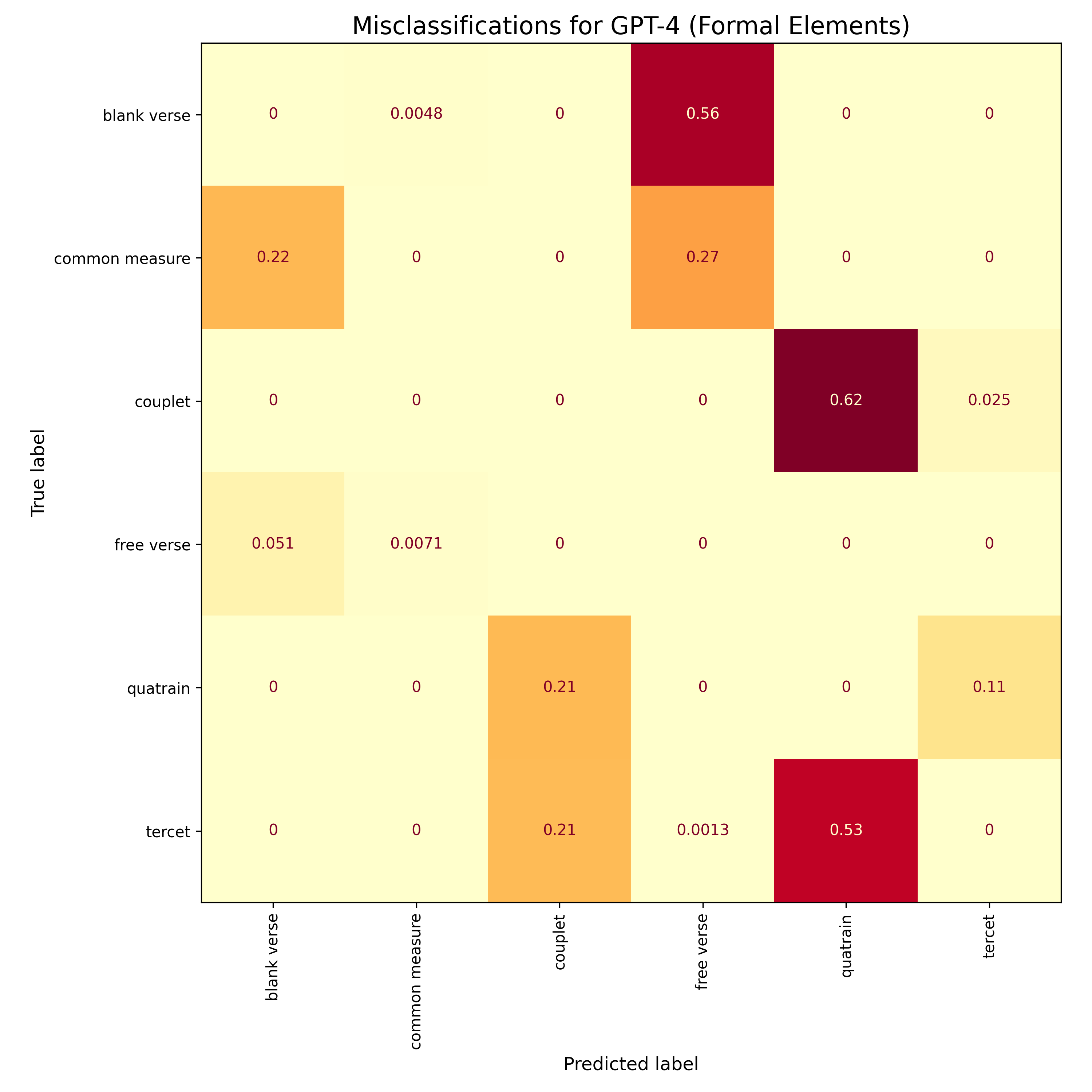}
    \caption{Misclassification proportions for formal elements by GPT-4.}
    \label{fig:mis-fixed-forms3}
\end{figure}

\clearpage

\subsection{Poetic Forms} 
\label{poetic-form-definitions}

Poetic forms can be defined and categorized in various ways. The definitions of forms and formal elements that we offer here are synthesized from information in glossaries of poetic terms available on the \href{https://www.poetryfoundation.org/learn/glossary-terms}{Poetry Foundation} and \href{https://poets.org/glossary}{Academy of American Poets} websites as well as from widely used poetry resources by \citet{strand_making_2000}, \citet{greene_princeton_2016}, and \citet{preminger_princeton_2015}. 

\subsubsection{Fixed Forms}

\paragraph{Ballad}\label{ballad} {\hyperref[table:evaluation-results]{$\hookleftarrow$}} A type of narrative poem with ties to music and oral performance. Traditional ballads often feature regular meter and stanzas. One conventional pattern is ``common measure,'' which consists of quatrains that rhyme ABCB and alternate iambic tetramater and trimeter.\\

Example Ballad: from ``\href{https://www.poetryfoundation.org/poems/50273/barbara-allen}{Barbara Allen}'' (by Anonymous)\\

In Scarlet town, where I was born,\\
\hspace*{.5cm}   There was a fair maid dwellin’,\\
Made every youth cry Well-a-way!\\
\hspace*{.5cm}   Her name was Barbara Allen.\\

All in the merry month of May,\\
\hspace*{.5cm}   When green buds they were swellin’,\\
Young Jemmy Grove on his death-bed lay,\\
\hspace*{.5cm}   For love of Barbara Allen.\\

He sent his man in to her then,\\
\hspace*{.5cm}   To the town where she was dwellin’;\\
“O haste and come to my master dear,\\
\hspace*{.5cm}   If your name be Barbara Allen...\\

\paragraph{Ghazal}\label{ghazal} {\hyperref[table:evaluation-results]{$\hookleftarrow$}} Originally an Arabic verse form, ghazals consists of a series of couplets usually all ending in the same word. Poets may include their name in the final couplet.\\

Example Ghazal: from ``\href{https://www.poetryfoundation.org/poetrymagazine/poems/144612/where-did-the-handsome-beloved-go}{Where did the handsome beloved go?}'' (by Jalal Al-Din Rumi, translated by Brad Gooch and Maryam Mortaz)\\

Where did the handsome beloved go?\\
I wonder, where did that tall, shapely cypress tree go?\\

He spread his light among us like a candle.\\
Where did he go? So strange, where did he go without me?\\

All day long my heart trembles like a leaf.\\
All alone at midnight, where did that beloved go?\\

Go to the road, and ask any passing traveler —\\ 
That soul-stirring companion, where did he go?\\

Go to the garden, and ask the gardener —\\
That tall, shapely rose stem, where did he go?\\

Go to the rooftop, and ask the watchman —\\ 
That unique sultan, where did he go?\\

\paragraph{Haiku}\label{haiku} {\hyperref[table:evaluation-results]{$\hookleftarrow$}} Originating in Japan, haiku are concise, non-narrative poems that often focus on imagery. In English, haiku often consist of three unrhymed lines with 5, 7, and 5 syllables respectively.\\

\noindent
Example Haiku: ``\href{https://www.poetryfoundation.org/poems/48708/in-kyoto-}{In Kyoto}'' (by Bashō, translated by Jane Hirshfield)\\

\begin{quote}
In Kyoto,\\
hearing the cuckoo,\\
I long for Kyoto.\\
\end{quote}

\paragraph{Limerick}\label{limerick} {\hyperref[table:evaluation-results]{$\hookleftarrow$}} A light, often comedic verse form consisting of five lines rhymed AABBA. In traditional limericks, lines 1, 2, and 5 are trimeter, while lines 3 and 4 are dimeter, and the dominant meter is anapestic.\\ 

Example Limerick: ``\href{https://www.poetryfoundation.org/poems/42910/a-young-lady-of-lynn}{A Young Lady of Lynn}'' (by Anonymous)\\

\begin{quote}
There was a young lady of Lynn,\\
Who was so uncommonly thin\\
    That when she essayed\\
    To drink lemonade\\
She slipped through the straw and fell in.\\
\end{quote}

\paragraph{Pantoum}\label{pantoum} {\hyperref[table:evaluation-results]{$\hookleftarrow$}} A Malaysian verse form that was adapted into French and later English, which consists of a series of quatrains in which the second and fourth lines of each quatrain serve as the first and third lines of the next quatrain. Pantoums do not have a determined length.\\ 

Example Pantoum: from ``\href{https://poets.org/poem/nocturne-5}{Nocturne}'' (by Sadakichi Hartmann)\\

\begin{quote}
    
Upon the silent sea-swept land\\
\hspace*{.5cm}The dreams of night fall soft and gray,\\
\hspace*{1cm}The waves fade on the jeweled sand\\
\hspace*{1.5cm}Like some lost hope of yesterday.\\

The dreams of night fall soft and gray\\
\hspace*{.5cm}Upon the summer-colored seas,\\
\hspace*{1cm}Like some lost hope of yesterday,\\
\hspace*{1.5cm}The sea-mew’s song is on the breeze.\\

Upon the summer-colored seas\\
\hspace*{.5cm}Sails gleam and glimmer ghostly white,\\
\hspace*{1cm}The sea-mew’s song is on the breeze\\
\hspace*{1.5cm}Lost in the monotone of night.\\

Sails gleam and glimmer ghostly white,\\
\hspace*{.5cm}They come and slowly drift away,\\
\hspace*{1cm}Lost in the monotone of night,\\
\hspace*{1.5cm}Like visions of a summer-day.\\

They shift and slowly drift away\\
\hspace*{.5cm}Like lovers’ lays that wax and wane,\\
\hspace*{1cm}The visions of a summer-day\\
\hspace*{1.5cm}Whose dreams we ne’er will dream again.\\
\end{quote}

\paragraph{Sestina}\label{sestina} {\hyperref[table:evaluation-results]{$\hookleftarrow$}} A complex verse form consisting of six, unrhymed, six-line stanzas followed by a three-line envoi. Each sestet includes the same six endwords in shifting, but specific patterns (below), and all six endwords also appear in the envoi. Endword pattern:\\
\begin{quote}
1: ABCDEF \\
2: FAEBDC \\
3: CFDABE \\
4: ECBFAD \\
5; DEACFB \\
6: BDFECA \\
envoi : ECA or  ACE\\
\end{quote}
Example Sestina: from ``\href{https://poets.org/poem/sestina-altaforte}{Sestina: Altaforte}'' (by Ezra Pound)\\

I\\
Damn it all!  all this our South stinks peace.\\
You whoreson dog, Papiols, come!  Let's to music!\\
I have no life save when the swords clash.\\
But ah!  when I see the standards gold, vair, purple, opposing\\
And the broad fields beneath them turn crimson,\\
Then howl I my heart nigh mad with rejoicing.\\

II\\
In hot summer have I great rejoicing\\
When the tempests kill the earth's foul peace,\\
And the lightnings from black heav'n flash crimson,\\
And the fierce thunders roar me their music\\
And the winds shriek through the clouds mad, opposing,\\
And through all the riven skies God's swords clash.\\

III\\
Hell grant soon we hear again the swords clash!\\
And the shrill neighs of destriers in battle rejoicing,\\
Spiked breast to spiked breast opposing!\\
Better one hour's stour than a year's peace\\
With fat boards, bawds, wine and frail music!\\
Bah!  there's no wine like the blood's crimson!\\

IV\\
And I love to see the sun rise blood-crimson.\\
And I watch his spears through the dark clash\\
And it fills all my heart with rejoicing\\
And pries wide my mouth with fast music\\
When I see him so scorn and defy peace,\\
His lone might 'gainst all darkness opposing.\\

V\\
The man who fears war and squats opposing\\
My words for stour, hath no blood of crimson\\
But is fit only to rot in womanish peace\\
Far from where worth's won and the swords clash\\
For the death of such sluts I go rejoicing;\\
Yea, I fill all the air with my music.\\

VI\\
Papiols, Papiols, to the music!\\
There's no sound like to swords swords opposing,\\
No cry like the battle's rejoicing\\
When our elbows and swords drip the crimson\\
And our charges 'gainst ``The Leopard's'' rush clash.\\
May God damn for ever all who cry ``Peace!''\\

VII\\
And let the music of the swords make them crimson!\\
Hell grant soon we hear again the swords clash!\\
Hell blot black for always the thought ``Peace!''\\

\paragraph{Sonnet}\label{sonnet} {\hyperref[table:evaluation-results]{$\hookleftarrow$}} A fourteen-line verse form, usually in iambic pentameter, and usually following a set rhyme scheme. The most common types of sonnets are Shakespearean/English, which consist of three quatrains followed by a couplet and often rhyme ABABCDCDEFEFGG, and Petrarchan/Italian, which consists of an octave followed by a sestet and often rhyme ABBAABBACDCDCD or ABBAABBACDECDE.\\

Example Petrarchan sonnet: ``\href{https://www.poetryfoundation.org/poems/44750/sonnet-19-when-i-consider-how-my-light-is-spent}{When I consider how my light is spent}'' (John Milton)\\

When I consider how my light is spent,\\
   Ere half my days, in this dark world and wide,\\
   And that one Talent which is death to hide\\
   Lodged with me useless, though my Soul more bent\\
To serve therewith my Maker, and present\\
   My true account, lest he returning chide;\\
   ``Doth God exact day-labour, light denied?''\\
   I fondly ask. But patience, to prevent\\
That murmur, soon replies, “God doth not need\\
   Either man’s work or his own gifts; who best\\
   Bear his mild yoke, they serve him best. His state\\
Is Kingly. Thousands at his bidding speed\\
   And post o’er Land and Ocean without rest:\\
   They also serve who only stand and wait.”\\

Example Shakespearean Sonnet: ``\href{https://www.poetryfoundation.org/poems/44691/america-56d223e1ac025}{America}'' (Claude McKay)\\

Although she feeds me bread of bitterness,\\
And sinks into my throat her tiger’s tooth,\\
Stealing my breath of life, I will confess\\
I love this cultured hell that tests my youth.\\
Her vigor flows like tides into my blood,\\
Giving me strength erect against her hate,\\
Her bigness sweeps my being like a flood.\\
Yet, as a rebel fronts a king in state,\\
I stand within her walls with not a shred\\
Of terror, malice, not a word of jeer.\\
Darkly I gaze into the days ahead,\\
And see her might and granite wonders there,\\
Beneath the touch of Time’s unerring hand,\\
Like priceless treasures sinking in the sand.\\

\paragraph{Villanelle}\label{villanelle} {\hyperref[table:evaluation-results]{$\hookleftarrow$}} A 19-line verse form originating in France, made up of five tercets followed by a quatrain, in which the first and third line of the first stanza are alternatingly repeated as a refrain in the following stanzas. Stanza 1 line 1 repeats as the third line of stanzas 2 and 4, and stanza 1 line 3 repeats as the third line of stanzas 3 and 5. These two lines also appear as the closing lines of the quatrain.\\

Example Villanelle: ``\href{https://poets.org/poem/do-not-go-gentle-good-night}{Do not go gentle into that good night}'' (Dylan Thomas)\\

Do not go gentle into that good night,\\
Old age should burn and rave at close of day;\\
Rage, rage against the dying of the light.\\

Though wise men at their end know dark is right,\\
Because their words had forked no lightning they\\
Do not go gentle into that good night.\\

Good men, the last wave by, crying how bright\\
Their frail deeds might have danced in a green bay,\\
Rage, rage against the dying of the light.\\

Wild men who caught and sang the sun in flight,\\
And learn, too late, they grieved it on its way,\\
Do not go gentle into that good night.\\

Grave men, near death, who see with blinding sight\\
Blind eyes could blaze like meteors and be gay,\\
Rage, rage against the dying of the light.\\

And you, my father, there on the sad height,\\
Curse, bless, me now with your fierce tears, I pray.\\
Do not go gentle into that good night.\\
Rage, rage against the dying of the light.\\

\subsubsection{Stanza Forms}

\paragraph{Couplet}\label{couplet} {\hyperref[table:memorization]{$\hookleftarrow$}} A two-line stanza or two lines of verse, often but not always rhymed.\\

Example Couplets: ``\href{https://www.poetryfoundation.org/poems/44830/interview-56d22412c4b44}{Interview}'' by Dorothy Parker\\

The ladies men admire, I’ve heard,\\
Would shudder at a wicked word.\\
Their candle gives a single light;\\
They’d rather stay at home at night.\\
They do not keep awake till three,\\
Nor read erotic poetry.\\
They never sanction the impure,\\
Nor recognize an overture.\\
They shrink from powders and from paints ...\\
So far, I’ve had no complaints.\\

\paragraph{Tercet}\label{tercet} {\hyperref[table:stanzas-meters-results]{$\hookleftarrow$}} A three-line stanza or three lines of verse, often but not always containing a rhyme.\\

Example Tercets: from ``\href{https://www.poetryfoundation.org/poems/47266/the-convergence-of-the-twain}{The Convergence of the Twain}'' (Thomas Hardy)\\

\textit{(Lines on the loss of the ``Titanic'')}\\
\begin{quote}
I\\
            In a solitude of the sea\\
            Deep from human vanity,\\
And the Pride of Life that planned her, stilly couches she.\\

II\\
            Steel chambers, late the pyres\\
            Of her salamandrine fires,\\
Cold currents thrid, and turn to rhythmic tidal lyres.\\

III\\
            Over the mirrors meant\\
            To glass the opulent\\
The sea-worm crawls — grotesque, slimed, dumb, indifferent.\\

IV\\
            Jewels in joy designed\\
            To ravish the sensuous mind\\
Lie lightless, all their sparkles bleared and black and blind.\\

V\\
            Dim moon-eyed fishes near\\
            Gaze at the gilded gear\\
And query: ``What does this vaingloriousness down here?'' ...\\ 
\end{quote}

\paragraph{Quatrain}\label{quatrain} {\hyperref[table:stanzas-meters-results]{$\hookleftarrow$}} A four-line stanza or unit of verse, often, but not always containing rhyme.\\

Example Quatrains: from ``\href{https://www.poetryfoundation.org/poems/44299/elegy-written-in-a-country-churchyard}{Elegy Written in a Country Churchyard}'' (Thomas Gray)\\

The curfew tolls the knell of parting day,\\
         The lowing herd wind slowly o'er the lea,\\
The plowman homeward plods his weary way,\\
         And leaves the world to darkness and to me.\\

Now fades the glimm'ring landscape on the sight,\\
         And all the air a solemn stillness holds,\\
Save where the beetle wheels his droning flight,\\
         And drowsy tinklings lull the distant folds; ...\\

\subsubsection{Meters}

\paragraph{Free Verse}\label{free-verse} {\hyperref[table:memorization]{$\hookleftarrow$}} Verse that does not follow a particular pattern of meter or rhyme.\\

Example Free Verse: from ``\href{https://www.poetryfoundation.org/poems/47311/the-waste-land}{The Waste Land}'' (T.S. Eliot)\\

April is the cruellest month, breeding\\
Lilacs out of the dead land, mixing\\
Memory and desire, stirring\\
Dull roots with spring rain.\\
Winter kept us warm, covering\\
Earth in forgetful snow, feeding\\
A little life with dried tubers.\\
Summer surprised us, coming over the Starnbergersee\\
With a shower of rain; we stopped in the colonnade,\\
And went on in sunlight, into the Hofgarten,\\
And drank coffee, and talked for an hour.\\
Bin gar keine Russin, stamm’ aus Litauen, echt deutsch.\\
And when we were children, staying at the archduke’s,\\
My cousin’s, he took me out on a sled,\\
And I was frightened. He said, Marie,\\
Marie, hold on tight. And down we went.\\
In the mountains, there you feel free.\\
I read, much of the night, and go south in the winter. ...\\

\paragraph{Blank Verse}\label{blank-verse} {\hyperref[table:memorization]{$\hookleftarrow$}} Unrhymed iambic pentameter.\\

Example Blank Verse: from \textit{\href{https://www.poetryfoundation.org/poems/45718/paradise-lost-book-1-1674-version}{Paradise Lost}} (John Milton)\\

Of Mans First Disobedience, and the Fruit\\
Of that Forbidden Tree, whose mortal tast\\
Brought Death into the World, and all our woe,\\
With loss of Eden, till one greater Man\\
Restore us, and regain the blissful Seat,\\
Sing Heav'nly Muse, that on the secret top\\
Of Oreb, or of Sinai, didst inspire\\
That Shepherd, who first taught the chosen Seed,\\
In the Beginning how the Heav'ns and Earth\\
Rose out of Chaos: or if Sion Hill\\
Delight thee more, and Siloa's brook that flow'd\\
Fast by the Oracle of God; I thence\\
Invoke thy aid to my adventrous Song,\\
That with no middle flight intends to soar\\
Above th' Aonian Mount, while it pursues\\
Things unattempted yet in Prose or Rhime.\\

\paragraph{Common Measure}\label{common-measure} {\hyperref[table:stanzas-meters-results]{$\hookleftarrow$}} Quatrains consisting of alternating lines of iambic tetrameter and trimeter, rhymed ABAB.\\

Example Common Measure: from ``\href{https://www.poetryfoundation.org/poems/44085/it-was-not-death-for-i-stood-up-355}{It was not death for I stood up}'' (Emily Dickinson)\\

\begin{quote}
It was not Death, for I stood up,\\
And all the Dead, lie down -\\
It was not Night, for all the Bells\\
Put out their Tongues, for Noon.\\

It was not Frost, for on my Flesh\\
I felt Siroccos - crawl -\\
Nor Fire - for just my marble feet\\
Could keep a Chancel, cool -\\

And yet, it tasted, like them all,\\
The Figures I have seen\\
Set orderly, for Burial\\
Reminded me, of mine -   ...\\
\end{quote}

\subsubsection{Unfixed forms}

\paragraph{Ode}\label{ode} {\hyperref[table:unfixed-forms-performance]{$\hookleftarrow$}} A formal lyric poem, which addresses or celebrates a person, place, object, or concept, usually that is not present. Odes are often longer verse forms, and their stanza patterns vary.\\ 

Example Ode: from ``\href{https://www.poetryfoundation.org/poems/44477/ode-on-a-grecian-urn}{Ode on a Grecian Urn}'' (John Keats)\\

Thou still unravish'd bride of quietness,\\
       Thou foster-child of silence and slow time,\\
Sylvan historian, who canst thus express\\
       A flowery tale more sweetly than our rhyme:\\
What leaf-fring'd legend haunts about thy shape\\
       Of deities or mortals, or of both,\\
               In Tempe or the dales of Arcady?\\
       What men or gods are these? What maidens loth?\\
What mad pursuit? What struggle to escape?\\
               What pipes and timbrels? What wild ecstasy?\\

Heard melodies are sweet, but those unheard\\
       Are sweeter; therefore, ye soft pipes, play on;\\
Not to the sensual ear, but, more endear'd,\\
       Pipe to the spirit ditties of no tone:\\
Fair youth, beneath the trees, thou canst not leave\\
       Thy song, nor ever can those trees be bare;\\
               Bold Lover, never, never canst thou kiss,\\
Though winning near the goal yet, do not grieve;\\
       She cannot fade, though thou hast not thy bliss,\\
               For ever wilt thou love, and she be fair!\\

\paragraph{Pastoral}\label{pastoral} {\hyperref[table:unfixed-forms-performance]{$\hookleftarrow$}} A type of poetry and a broader creative tradition idealizing rural life. \\

Example Pastoral: ``\href{https://www.poetryfoundation.org/poems/44675/the-passionate-shepherd-to-his-love}{The Passionate Shepherd to His Love}'' (Christopher Marlowe)\\
\begin{quote}

Come live with me and be my love,\\
And we will all the pleasures prove,\\
That Valleys, groves, hills, and fields,\\
Woods, or steepy mountain yields.\\

And we will sit upon the Rocks,\\
Seeing the Shepherds feed their flocks,\\
By shallow Rivers to whose falls\\
Melodious birds sing Madrigals.\\

And I will make thee beds of Roses\\
And a thousand fragrant posies,\\
A cap of flowers, and a kirtle\\
Embroidered all with leaves of Myrtle;\\

A gown made of the finest wool\\
Which from our pretty Lambs we pull;\\
Fair lined slippers for the cold,\\
With buckles of the purest gold;\\

A belt of straw and Ivy buds,\\
With Coral clasps and Amber studs:\\
And if these pleasures may thee move,\\
Come live with me, and be my love.\\

The Shepherds’ Swains shall dance and sing\\
For thy delight each May-morning:\\
If these delights thy mind may move,\\
Then live with me, and be my love.\\
\end{quote}

\paragraph{Aubade}\label{aubade} {\hyperref[table:unfixed-forms-performance]{$\hookleftarrow$}} A poem or song welcoming or lamenting the arrival of dawn, usually with romantic themes.\\

Example Aubade: ``\href{https://www.poetryfoundation.org/poems/51783/break-of-day}{Break of Day}'' (John Donne)\\

‘Tis true, ‘tis day, what though it be?\\
O wilt thou therefore rise from me?\\
Why should we rise because ‘tis light?\\
Did we lie down because ‘twas night?\\
Love, which in spite of darkness brought us hither,\\
Should in despite of light keep us together.\\

Light hath no tongue, but is all eye;\\
If it could speak as well as spy,\\
This were the worst that it could say,\\
That being well I fain would stay,\\
And that I loved my heart and honour so,\\
That I would not from him, that had them, go.\\

Must business thee from hence remove?\\
Oh, that’s the worst disease of love,\\
The poor, the foul, the false, love can\\
Admit, but not the busied man.\\
He which hath business, and makes love, doth do\\
Such wrong, as when a married man doth woo.\\

\paragraph{Dramatic Monologue}\label{dramatic} {\hyperref[table:memorization]{$\hookleftarrow$}} A poem in which a usually fictional speaker addresses a listener, who is also often imagined. \\

Example Dramatic Monologue: from ``\href{https://www.poetryfoundation.org/poems/43768/my-last-duchess}{My Last Duchess}'' (Robert Browning)\\

That’s my last Duchess painted on the wall,\\
Looking as if she were alive. I call\\
That piece a wonder, now; Fra Pandolf’s hands\\
Worked busily a day, and there she stands.\\
Will’t please you sit and look at her? I said\\
“Fra Pandolf” by design, for never read\\
Strangers like you that pictured countenance,\\
The depth and passion of its earnest glance,\\
But to myself they turned (since none puts by\\
The curtain I have drawn for you, but I)\\
And seemed as they would ask me, if they durst,\\
How such a glance came there; so, not the first\\
Are you to turn and ask thus. Sir, ’twas not\\
Her husband’s presence only, called that spot\\
Of joy into the Duchess’ cheek; perhaps\\
Fra Pandolf chanced to say, “Her mantle laps ...\\

\paragraph{Elegy}\label{elegy} {\hyperref[table:memorization]{$\hookleftarrow$}} A form of poetry and broader mode of writing expressing grief or loss, often in relation to its subject’s death.\\

Example Elegy: from ``\href{https://www.poetryfoundation.org/poems/44733/lycidas}{Lycidas}'' (John Milton)\\

Yet once more, O ye laurels, and once more\\
Ye myrtles brown, with ivy never sere,\\
I come to pluck your berries harsh and crude,\\
And with forc'd fingers rude\\
Shatter your leaves before the mellowing year.\\
Bitter constraint and sad occasion dear\\
Compels me to disturb your season due;\\
For Lycidas is dead, dead ere his prime,\\
Young Lycidas, and hath not left his peer.\\
Who would not sing for Lycidas? he knew\\
Himself to sing, and build the lofty rhyme.\\
He must not float upon his wat'ry bier\\
Unwept, and welter to the parching wind,\\
Without the meed of some melodious tear.\\

\paragraph{Concrete Poetry}\label{concrete} {\hyperref[table:unfixed-forms-performance]{$\hookleftarrow$}} A type of poetry that is structured by visual effect on the page, and often emphasizes nonlinguistic aspects of writing, including typography, layout, whitespace, etc.\\

Example Concrete Poetry: ``\href{https://www.poetryfoundation.org/poems/44361/easter-wings}{Easter Wings}'' (George Herbert)\\

Lord, who createdst man in wealth and store,\\
      \hspace*{.5cm}Though foolishly he lost the same,\\
             \hspace*{1cm}Decaying more and more,\\
                   \hspace*{1.5cm}Till he became\\
                        \hspace*{2cm}Most poore:\\
                        \hspace*{2cm}With thee\\
                  \hspace*{1.5cm}O let me rise\\
            \hspace*{1cm}As larks, harmoniously,\\
      \hspace*{.5cm}And sing this day thy victories:\\
Then shall the fall further the flight in me.\\

My tender age in sorrow did beginne\\
      \hspace*{.5cm}And still with sicknesses and shame.\\
            \hspace*{1cm}Thou didst so punish sinne,\\
                  \hspace*{1.5cm}That I became\\
                        \hspace*{2cm}Most thinne.\\
                        \hspace*{2cm}With thee\\
                  \hspace*{1.5cm}Let me combine,\\
            \hspace*{1cm}And feel thy victorie:\\
         \hspace*{1cm}For, if I imp my wing on thine,\\
\hspace*{.5cm}Affliction shall advance the flight in me.\\

\paragraph{Prose Poem}\label{prose} {\hyperref[table:memorization]{$\hookleftarrow$}}  A poetic composition that is not broken up into lines.\\

Example Prose Poem: \textit{\href{https://poets.org/poem/gitanjali-14}{Gitanjali}}, \#14 (by Rabindranath Tagore)\\

My desires are many and my cry is pitiful, but ever didst thou save me by hard refusals; and this strong mercy has been wrought into my life through and through.\\

Day by day thou art making me worthy of the simple, great gifts that thou gavest to me unasked—this sky and the light, this body and the life and the mind—saving me from perils of overmuch desire.\\

There are times when I languidly linger and times when I awaken and hurry in search of my goal; but cruelly thou hidest thyself from before me.\\

Day by day thou art making me worthy of thy full acceptance by refusing me ever and anon, saving me from perils of weak, uncertain desire.\\

\paragraph{Ars Poetica}\label{ars} {\hyperref[table:unfixed-forms-performance]{$\hookleftarrow$}} A poem about poetry.\\

Example Ars Poetica: from ``\href{https://poets.org/poem/poetry}{Poetry}'' (Marianne Moore)\\

I too, dislike it: there are things that are important beyond all this fiddle.\\
   Reading it, however, with a perfect contempt for it, one discovers that there is in\\
   it after all, a place for the genuine.\\
      Hands that can grasp, eyes\\
      that can dilate, hair that can rise\\
         if it must, these things are important not because a\\

high-sounding interpretation can be put upon them but because they are\\
   useful; when they become so derivative as to become unintelligible, the\\
   same thing may be said for all of us—that we\\
      do not admire what\\
      we cannot understand. The bat,\\
         holding on upside down or in quest of something to\\
         
eat, elephants pushing, a wild horse taking a roll, a tireless wolf under\\
   a tree, the immovable critic twinkling his skin like a horse that feels a flea, the base—\\
   ball fan, the statistician—case after case\\
      could be cited did\\
      one wish it; nor is it valid\\
         to discriminate against ``business documents and\\

school-books''; all these phenomena are important. One must make a distinction\\
   however: when dragged into prominence by half poets, the result is not poetry,\\
   nor till the autocrats among us can be\\
     ``literalists of\\
      the imagination”—above\\
         insolence and triviality and can present\\

for inspection, imaginary gardens with real toads in them, shall we have\\
   it. In the meantime, if you demand on the one hand, in defiance of their opinion—\\
   the raw material of poetry in\\
      all its rawness, and\\
      that which is on the other hand,\\
         genuine, then you are interested in poetry.\\

\paragraph{Ekphrasis}\label{ekphra} {\hyperref[table:memorization]{$\hookleftarrow$}} Writing that uses vivid language to respond to or describe a work of visual art. \\

Example Ekphrasis: ``\href{https://poets.org/poem/seeing-elgin-marbles}{On Seeing the Elgin Marbles}'' (John Keats)\\

My spirit is too weak—mortality\\
   Weighs heavily on me like unwilling sleep,\\
   And each imagined pinnacle and steep\\
Of godlike hardship tells me I must die\\
Like a sick eagle looking at the sky.\\
   Yet ‘tis a gentle luxury to weep,\\
   That I have not the cloudy winds to keep,\\
Fresh for the opening of the morning’s eye.\\
Such dim-conceived glories of the brain\\
   Bring round the heart an indescribable feud;\\
So do these wonders a most dizzy pain,\\
   That mingles Grecian grandeur with the rude\\
Wasting of old Time—with a billowy main—\\
   A sun—a shadow of a magnitude.
\end{document}